\pdfoutput=1

\documentclass[a4paper,fleqn]{cas-dc}
\usepackage[numbers, sort&compress]{natbib}
\usepackage{amsmath}
\usepackage{amssymb}
\usepackage{orcidlink}
\usepackage{caption}
\usepackage{bbding}
\usepackage{dsfont}
\usepackage{graphicx}
\usepackage{chngcntr}
\usepackage{float}
\usepackage{diagbox}
\usepackage{multirow}
\usepackage{booktabs}
\usepackage{textcomp}
\usepackage{subcaption}
\usepackage[title]{appendix}
\usepackage{stfloats}
\begin{document}

\let\WriteBookmarks\relax
\def\floatpagepagefraction{1}
\def\textpagefraction{.001}
\shorttitle{PMFSNet: Polarized Multi-scale Feature Self-attention Network For Lightweight Medical Image Segmentation}
\shortauthors{Jiahui Zhong et~al.}

\title [mode = title]{PMFSNet: Polarized Multi-scale Feature Self-attention Network For Lightweight Medical Image Segmentation} 

\author[1]{Jiahui Zhong}[orcid=0009-0002-3788-4471]
\ead{yykzhjh@std.uestc.edu.cn}
\credit{Conceptualization, Data curation, Investigation, Methodology, Project administration, Writing - original draft}

\author[1]{Wenhong Tian}
\ead{tian_wenhong@uestc.edu.cn}
\cormark[1]
\credit{Conceptualization, Funding acquisition, Methodology, Supervision, Validation, Writing - review \& editing}

\author[1]{Yuanlun Xie}[orcid=0000-0001-9682-2065]
\ead{fengyuxiexie@163.com}
\cormark[1]
\credit{Conceptualization, Methodology, Writing - review \& editing}

\author[1]{Zhijia Liu}
\ead{zhijialiu82@gmail.com}
\credit{Data curation, Writing - review \& editing}

\author[1]{Jie Ou}
\ead{oujieww6@gmail.com}
\credit{Writing - review \& editing}

\author[2]{Taoran Tian}
\ead{taoran.tian@scu.edu.cn}
\credit{Data curation, Writing - review \& editing}

\author[3]{Lei Zhang}
\ead{lzhang@lincoln.ac.uk}
\credit{Conceptualization, Writing - review \& editing}

\address[1]{School of Information and Software Engineering, University of Electronic Science and Technology of China, Chengdu, 610054, P. R. China}
\address[2]{State Key Laboratory of Oral Diseases, National Clinical Research Center for Oral Diseases, West China Hospital of Stomatology, Sichuan University, Chengdu, 610041, P. R. China}
\address[3]{School of Computer Science, University of Lincoln, LN6 7TS, UK}
\cortext[cor1]{Corresponding author}

\begin{abstract}
Current state-of-the-art medical image segmentation methods prioritize accuracy but often at the expense of increased computational demands and larger model sizes. Applying these large-scale models to the relatively limited scale of medical image datasets tends to induce redundant computation, complicating the process without the necessary benefits. This approach not only adds complexity but also presents challenges for the integration and deployment of lightweight models on edge devices. For instance, recent transformer-based models have excelled in 2D and 3D medical image segmentation due to their extensive receptive fields and high parameter count. However, their effectiveness comes with a risk of overfitting when applied to small datasets and often neglects the vital inductive biases of Convolutional Neural Networks (CNNs), essential for local feature representation. In this work, we propose PMFSNet, a novel medical imaging segmentation model that effectively balances global and local feature processing while avoiding the computational redundancy typical in larger models. PMFSNet streamlines the UNet-based hierarchical structure and simplifies the self-attention mechanism's computational complexity, making it suitable for lightweight applications. It incorporates a plug-and-play PMFS block, a multi-scale feature enhancement module based on attention mechanisms, to capture long-term dependencies. Extensive comprehensive results demonstrate that even with a model ($\leq$ 1 million parameters), our method achieves superior performance in various segmentation tasks across different data scales. It achieves (IoU) metrics of 84.68\%, 82.02\%, and 78.82\% on public datasets of teeth CT (CBCT), ovarian tumors ultrasound(MMOTU), and skin lesions dermoscopy images (ISIC 2018), respectively. The source code is available at \href{https://github.com/yykzjh/PMFSNet}{https://github.com/yykzjh/PMFSNet}.

\end{abstract}

\maketitle

\begin{sloppypar}
\section{Introduction}
\begin{figure*}
    \begin{subfigure}{0.407\linewidth}
        \flushleft
        \includegraphics[width=0.99\linewidth, height=6cm]{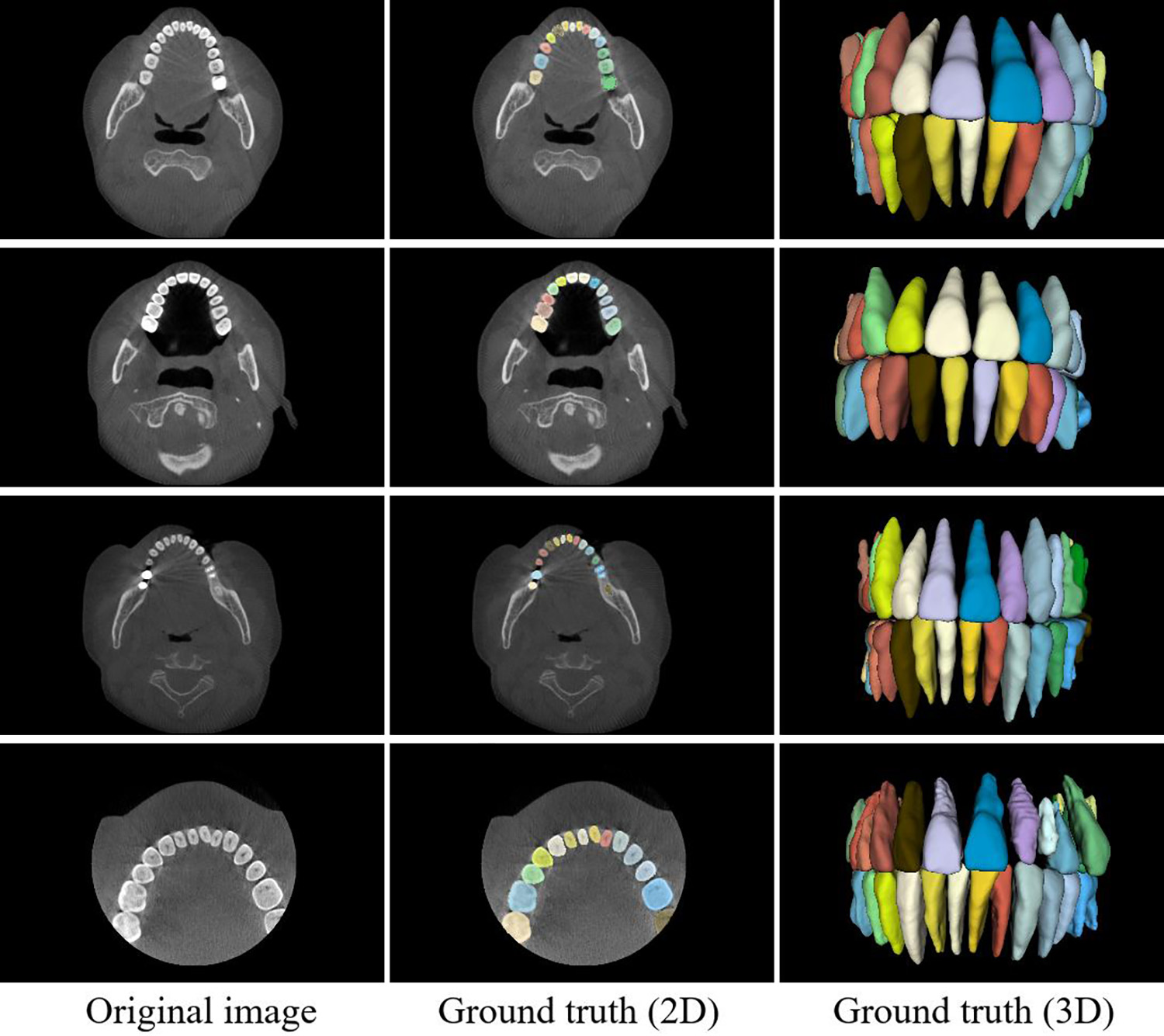} 
        \caption{The 3D CBCT tooth dataset samples.}
        \label{fig: 3D_CBCT_Tooth_samples}
    \end{subfigure}
    \begin{subfigure}{0.271\linewidth}
        \includegraphics[width=0.99\linewidth, height=6cm]{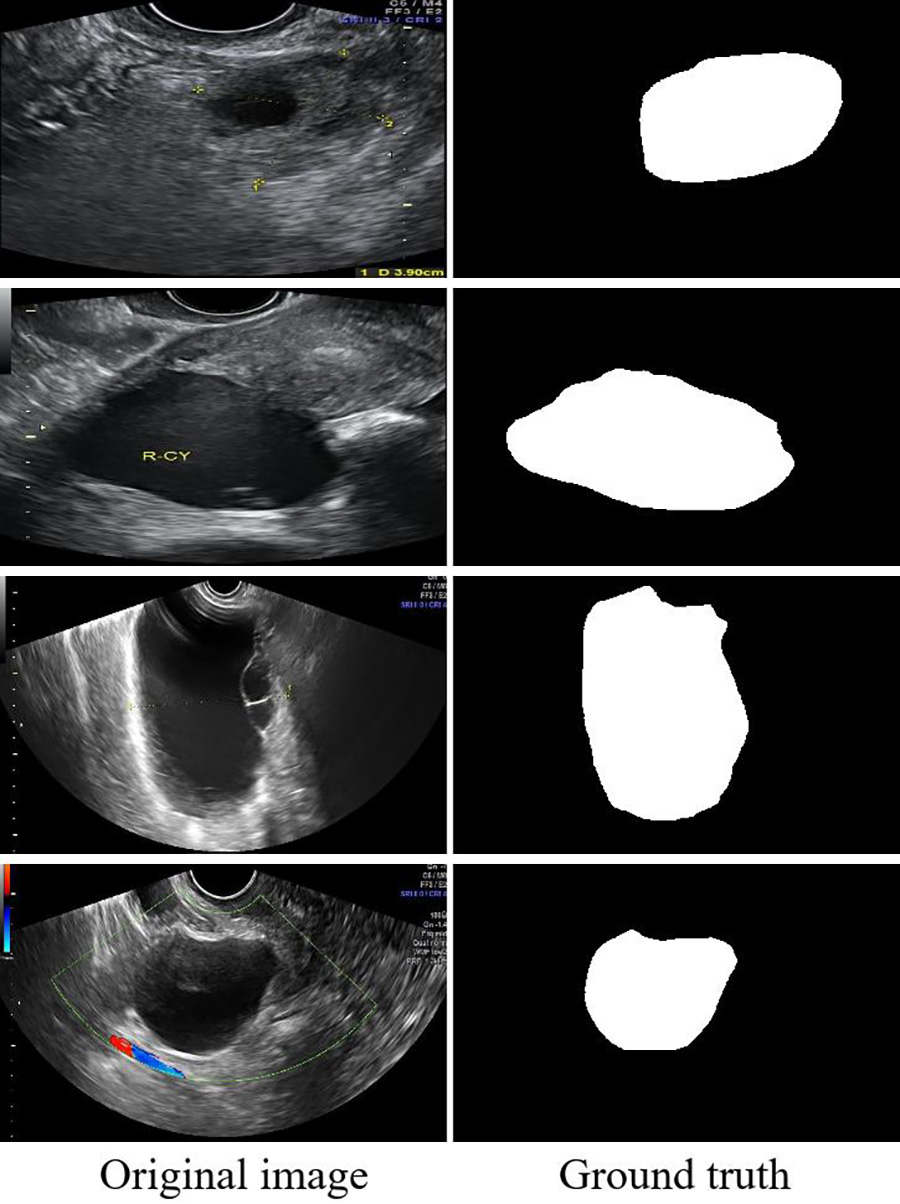}
        \caption{The MMOTU dataset samples.}
        \label{fig: MMOTU_samples}
    \end{subfigure}
    \begin{subfigure}{0.271\linewidth}
        \flushright
        \includegraphics[width=0.99\linewidth, height=6cm]{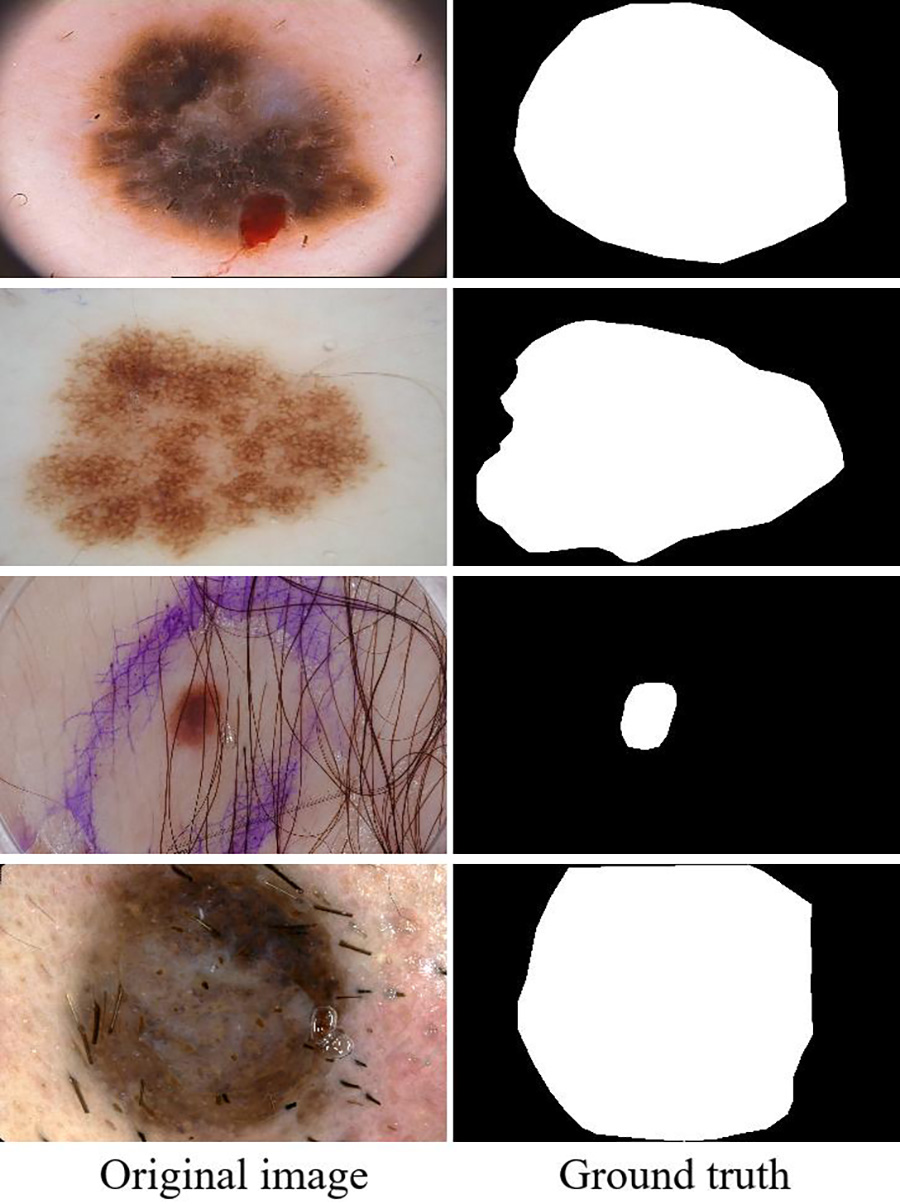}
        \caption{The ISIC 2018 dataset samples.}
        \label{fig: ISIC_2018_samples}
    \end{subfigure}
    \caption{The samples from different datasets. Sub-figure (a) shows some challenging samples of the 3D CBCT tooth dataset, such as missing teeth, metal artifacts, and incomplete views. Sub-figure (b) shows some challenging samples of the MMOTU dataset, such as inconspicuous lesion regions, blurred lesion boundaries, and low-contrast samples. Sub-figure (c) shows some challenging samples of the ISIC 2018 dataset, such as blurred samples, irregular lesion boundaries, and occluded lesions.}
    \label{fig: samples}
\end{figure*}

Medical image segmentation plays a key role in Computer Aided Diagnosis (CAD) systems as it is often the key step for the analysis of anatomical structures \cite{MIS}. Precise delineation of tissues and lesions plays a crucial role in quantifying diseases, aiding the assessment of disease prognosis, and evaluating treatment efficacy. However, several challenges remain open issues in the field of medical image segmentation, particularly due to the diversity of imaging modalities, data limitations, and the complexity of different clinical scenarios. More specifically, 1) Medical imaging encompasses a wide range of modalities, including Computed Tomography (CT), Magnetic Resonance Imaging (MRI), X-ray, and Ultrasound Imaging (UI). This diversity complicates the development of universal or adaptable algorithms. 2) Due to privacy concerns and the complexities involved in labeling, the size of medical images is relatively smaller than the natural images. Additionally, the process of collecting medical images with accurate ground truth annotations is labor-intensive and time-consuming which creates a significant barrier to the compilation of high-quality datasets. 3) Clinical practice often involves complicated cases that are challenging for image segmentation algorithms. These include scenarios with low-contrast lesions, irregular lesion shapes (Figure \ref{fig: ISIC_2018_samples}), blurred lesion boundaries (Figure \ref{fig: MMOTU_samples}), and metal artifacts in images (Figure \ref{fig: 3D_CBCT_Tooth_samples}). Such complexities can hinder the performance of segmentation models.

\begin{figure}
    \centering
	\includegraphics[width=0.99\linewidth]{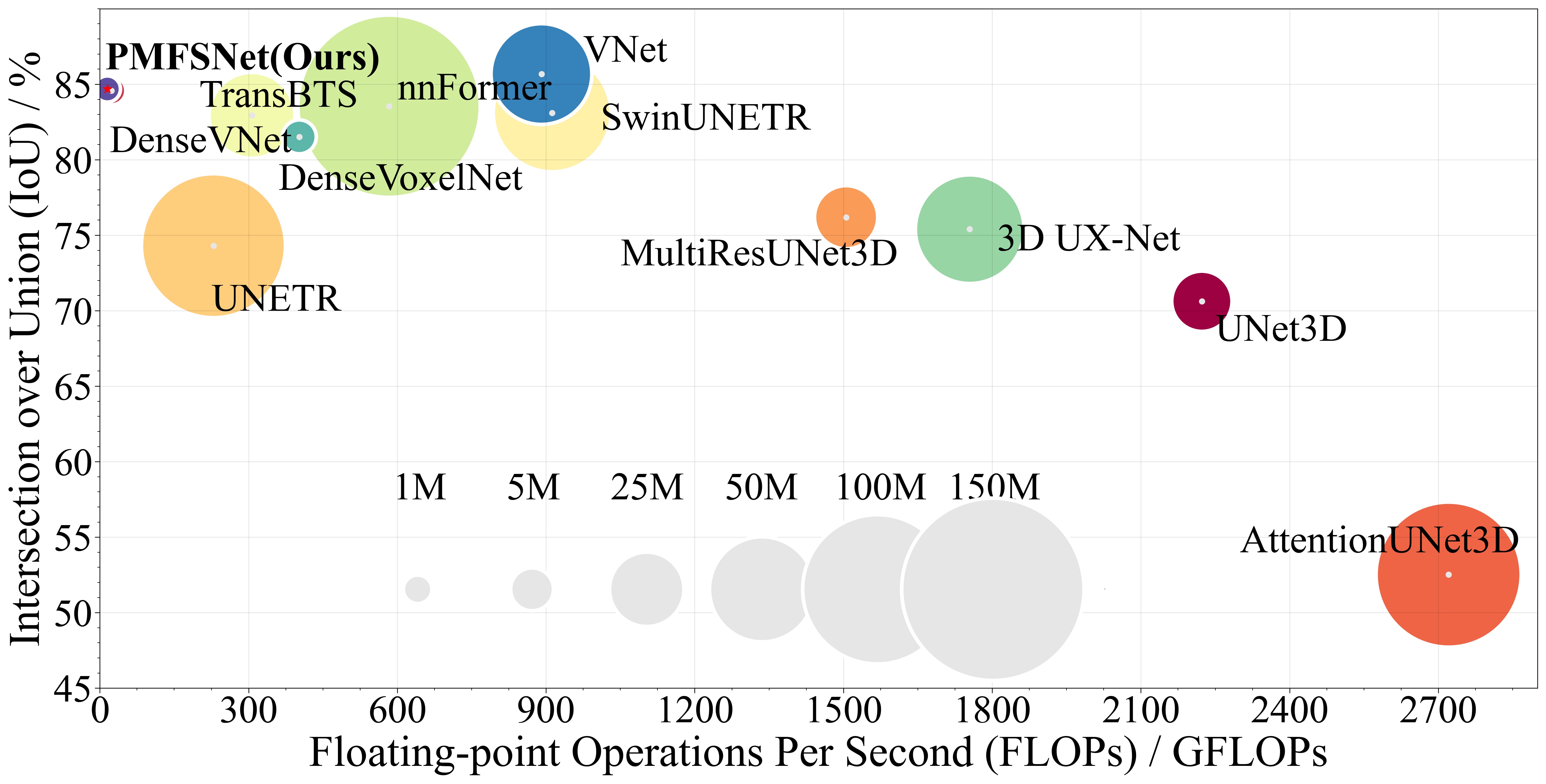}
	\caption{Params-FLOPs-IoU correlation comparison on the 3D CBCT tooth dataset. The Y-axis corresponds to the Intersection over Union (IoU) (higher is better), the X-axis corresponds to the Floating-point Operations Per Second (FLOPs) (lower is better), and the size of the circle corresponds to the Parameters (Params) (smaller is better). PMFSNet (ours) has achieved the best results in balance segmentation performance, model parameters count, and computational complexity.}
	\label{fig: 3D_CBCT_Tooth_bubble_image}
\end{figure}

Leveraging the significant advancements in deep learning, segmentation approaches employing Deep Convolutional Neural Networks (DCNNs) have demonstrated promising results \cite{FCN, PSPNet}. In particular, architectures based on UNet \cite{UNet} and its variants have achieved state-of-the-art results in various medical semantic segmentation tasks. UNet is based on an encoder-decoder architecture, which is hierarchical downsampled convolutional layers accompanied by symmetric upsampled convolutional layers. Furthermore, skip connections effectively bridge the semantic gap between encoders and decoders, via fusing initial features to the decoder. UNet++ \cite{UNet++} utilizes nested and dense skip connections to provide accurate semantic and coarse grading information to the decoder. R2U-Net \cite{R2UNet} is constructed based on the residual concept and recurrent technique. Attention U-Net \cite{AttentionUNet} highlights the salient features through skip connections and applies information extracted from coarse scales to gating, eliminating irrelevant and noisy responses generated by skip connections. In addition, to fully utilize the voxel information in CTs and MRIs, some models of 3D CNNs have been proposed \cite{UNet3D, VNet}.

Despite the robust representation learning abilities of methods based on UNet and its variants, their effectiveness in capturing long-term dependencies is constrained by their inherently local receptive fields \cite{Local_Relation_Networks, Stand-alone_self-attention}. Some studies have attempted to enlarge the receptive field with atrous convolution \cite{Deeplabv3+, CENet}. However, the locality of the receptive fields in convolutional layers still limits their learning capabilities to relatively small regions.

Recently, the Vision Transformers (ViTs) \cite{ViTs} have demonstrated excellent capabilities in representative learning, particularly in volumetric medical image segmentation. TransUNet \cite{TransUNet} further combines the functionality of ViT with the advantages of UNet in the field of medical image segmentation. CFATransUnet \cite{CFATransUnet} combines the channel-wise cross-fusion attention and transformer to integrate the channel dimension’s global features, strengthening the effective information and limiting the irrelevant features more strongly. MedT \cite{MedT} introduces additional control mechanisms in the self-attention module and proposes a local-global training strategy (LoGo) to further improve the performance. TransBTS \cite{TransBTS} is the first to utilize the transformer in 3D CNN for MRI brain tumor segmentation. UNETR \cite{UNETR} employs pure transformers as encoders to redesign the task of volumetric (3D) medical image segmentation as a sequence-to-sequence prediction problem, effectively capturing global multi-scale information. Swin-Unet \cite{SwinUNet} builds an encoder, bottleneck, and decoder based on the pure transformer.

However, while these transformers exhibit competitive performance, they involve training on substantial amounts of data and come with the cost of the heightened model complexity. This is because the computational burden of most existing transformers, which rely on self-attention operations, exhibits quadratic complexity. For example, These efforts \cite{SwinUNETR, nnFormer} have mainly focused on improving segmentation accuracy, which in turn greatly increases the size of the model in terms of Parameters (Params) and Floating-point Operations Per Second (FLOPs). Consequently, this leads to compromised robustness, posing challenges for system integration, particularly in embedded environments. In this case, finding ways to optimize the model design, eliminate redundant computations, and effectively utilize the existing parameters without compromising performance for medical image segmentation, remains a critical challenge.

In this work, our primary aim is to achieve the trade-off between performance and efficiency within a generalized, lightweight framework for medical image segmentation (Figure \ref{fig: 3D_CBCT_Tooth_bubble_image}). This framework is designed to be adaptable across various imaging modalities. Considering the issues and limitations mentioned above, we optimize the self-attention computation and propose the PMFSNet architecture. We introduce a polarized multi-scale feature self-attention (PMFS) block at the network's bottleneck to directly enhance global multi-scale low-dimensional features to encode long-term dependencies and enrich feature representations. The PMFS block decreases the computational complexity of self-attention from quadratic to linear by simplifying the computation of key vectors for each attention point to the computation of the global key vector. Meanwhile, to balance the performance, multi-scale features are introduced to expand the number of attention points and increase the multi-scale long-term dependencies.

Our main contributions are summarized as follows:

\begin{enumerate}
    \item[1)] We propose a lightweight PMFSNet model with $\leq$ 1 million parameters, designed for both 2D and 3D medical image segmentation.

    \item[2)] A plug-and-play global attention module is proposed that can improve segmentation by learning the long-term dependencies without significantly increasing the parameter count.

    \item[3)] Our proposed model exhibits competitive performance across three datasets, accomplishing this with significantly fewer model parameters compared to current state-of-the-art (SOTA) models.

    \item[4)] Our model's reduced complexity, without sacrificing performance, demonstrates its value in model integration and deployment, especially in resource-constrained environments.
\end{enumerate}

The rest of the paper is organized as follows. Section \ref{Related Work} presents a review of related works. Section \ref{Our Method} gives detailed information about the proposed PMFSNet. Section \ref{Experiments} shows the experiments and results. Section \ref{Discussion} is the discussion part. Followed by the conclusion in section \ref{Conclusion}.

\section{Related Work}\label{Related Work}
\subsection{CNNs-based Medical Image Segmentation}
Since the introduction of the seminal U-Net \cite{UNet}, CNN-based networks have achieved state-of-the-art results on various 2D and 3D medical image segmentation tasks. One of the challenges in medical image segmentation is the considerable variation in scales among different objects. To address this problem, UNet creates shortcuts between the encoder and decoder to capture contextual and precise positioning information. Zhou et al. \cite{UNet++} connect all U-Net layers (one through four) to form U-Net++. Liu et al. \cite{ResPath} introduce the Residual Path (ResPath) method, involving additional convolution operations on encoder features prior to their fusion with corresponding decoder features. Chen et al. \cite{Deeplabv3+} propose the atrous spatial pyramid pooling (ASPP) module and the well-known DeepLab family networks, which show strong recognition capability on the same objects at different scales. For volume-wise segmentation, 3D models directly utilize the full volumetric image represented by a sequence of 2D slices or modalities. To exploit 3D context and cope with computational resource limitations, Isensee et al. \cite{nnUNet} propose to extract features at multiple scales or assembled frameworks. However, a main limitation of these networks is their subpar performance in learning global context and long-term spatial dependencies, which can hinder segmentation effectiveness, particularly for those challenging cases (low-contrast lesions, irregular lesion shapes, blurred lesion boundaries, and metal artifacts, e.g. Figure \ref{fig: samples}).

\subsection{Lightweight Methods}
A series of methods \cite{GoogleNet, Xception, MobilenetV2, Dilated_Convolution, SqueezeNet} have been proposed to tackle the issue of high computational costs in models, focusing on enhancing feature representation in lightweight models or reducing their computational complexity without compromising performance. In GoogleNet, Szegedy et al. \cite{GoogleNet} propose the Inception architecture, which can obtain richer features without increasing network depth by merging convolution kernels in parallel rather than simply stacking convolution layers. In Xception, Chollet \cite{Xception} generalizes the ideas of separable convolutions in the Inception series and proposes a depthwise separable convolution, which can drastically reduce computation complexity by factorizing a standard convolution into a depthwise convolution followed by a pointwise convolution (i.e., 1 × 1 convolution). Thanks to the success of Xception, depthwise separable convolution has become an essential component of MobileNetV2 \cite{MobilenetV2}. The dilated (atrous) convolution \cite{Dilated_Convolution} introduces ordinary convolution layers with the parameter of dilation rate to enlarge the receptive field size without increasing the number of training parameters. The concept of group convolution is first proposed in SqueezeNet \cite{SqueezeNet}, and it is further successfully adopted in ResNeXt \cite{ResNeXt}. However, standard group convolution does not communicate between different groups, which restricts their representation capability. To solve this problem, ShuffleNet \cite{ShuffleNet} proposes a channel shuffle unit to randomly permute the output channels of group convolution to make the features of different groups fully intercourse.

Despite these advances in lightweight model development, their adoption in 3D volume segmentation remains limited. In contrast, current state-of-the-art methods for medical image segmentation mainly focus on enhancing accuracy, frequently resulting in higher computational costs. This trend has inspired us to design a lightweight segmentation model specifically aimed at reducing redundant computations during the learning process, with a strong emphasis on thoroughly investigating and optimizing the efficiency of parameter usage.

\subsection{Attention Mechanisms and Transformers}
Attention mechanisms are essential for directing model focus towards relevant features, enhancing performance. In recent years, the attention mechanism substantially promotes semantic segmentation. Oktay et al. \cite{AttentionUNet} propose an attention U-Net. Before fusing the feature maps from the encoder and decoder, attention U-Net inserts an attention gate to control the spatial feature importance. Besides applying the spatial attention mechanism, Chen et al. \cite{FEDNet} propose FED-Net, which uses a channel attention mechanism to improve the performance of liver lesion segmentation. There are also some notable works \cite{FocusNet, NonlocalUNet} aiming at mixing spatial and channel attention mechanisms. Transformer with Multiple Head Self-Attention (MHSA) has also made significant progress in medical image segmentation tasks with the introduction of Vision Transformers (ViTs) \cite{ViTs}. Significant efforts have been put into integrating ViTs for dense predictions in the medical imaging domain \cite{UNETR, TransUNet, nnFormer, TransBTS}. With the advancement of the Swin Transformer, SwinUNETR \cite{SwinUNETR} equips the encoder with the Swin Transformer blocks to compute self-attention for enhancing brain tumor segmentation accuracy in 3D MRI Images. Another Unet-like architecture Swin-Unet \cite{SwinUNet} further adopts the Swin Transformer on both the encoder and decoder network via skip-connections to learn local and global semantic features for multi-abdominal CT segmentation. The SwinBTS \cite{SwinBTS} has a similar intrinsic structure to Swin-Unet with an enhanced transformer module for detailed feature extraction. Nevertheless, transformer-based frameworks for volumetric segmentation still necessitate extended training periods and involve high computational complexity, particularly when extracting features across multiple scales \cite{Cotr, Transformer_Survey}.

Several studies \cite{LeaNet, LM-Net} have explored integrating attention modules or Vision Transformers (ViTs) into lightweight architectures. For instance, LeaNet \cite{LeaNet} utilizes various attention modules for edge segmentation. LM-Net \cite{LM-Net} optimizes transformer modules to address the challenge of ViTs needing extensive pre-training on large datasets. However, these models are tailored for lightweight processing in 2D image modalities. Our model is versatile, designed to accommodate both 2D and 3D medical imaging modalities, and features a more flexible attention module.

\section{Our Method}\label{Our Method}

\begin{figure*}
    \centering
	\includegraphics[width=\textwidth]{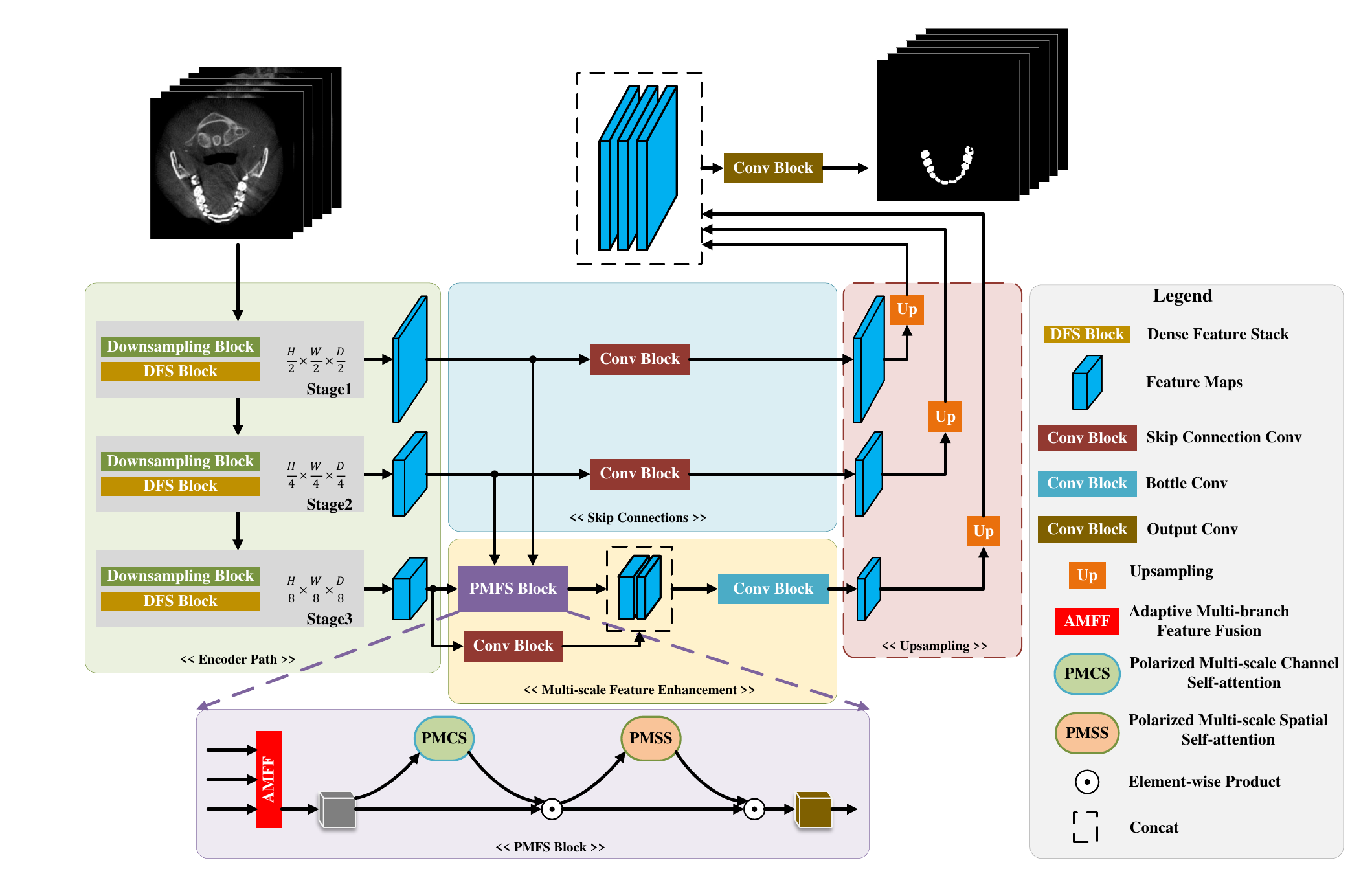}
	\caption{Overview of Polarized Multi-scale Feature Self-attention Network (PMFSNet) architecture. The input medical images are fed into an encoder with 3 stages. Then, the PMFS block enhances the features of the network's bottleneck using features at different scales. Finally, the skip connections are fused with an optional decoder, which sequentially incorporates the global contextual features into the enhanced bottleneck features by CNN-based up-sampling, gradually restoring them to the same resolution as the input image.}
	\label{fig: PMFSNet}
\end{figure*}

\subsection{Overview of PMFSNet} 
The overall architecture of the PMFSNet is illustrated in Figure \ref{fig: PMFSNet}. The model consists of four core components: an encoder, skip connections, a Polarized Multi-scale Feature Self-attention (PMFS) block, and an optional decoder. In particular, each encoder stage utilizes dense-feature-stacking convolution to retain as much multi-scale semantic information as possible. We have designed the decoder to be optional, and adaptable to tasks of varying magnitudes. Meanwhile, the architecture employs skip connections to capture contextual information and perform multi-scale feature extraction.

The PMFSNet enhances the fusion of multi-scale semantic information and global contextual features during the learning process, where the PMFS block enables more nuanced and precise encoding of multi-scale long-term dependencies. The PMFS block is also in turn designed to efficiently compensate for any performance loss that may result from any adaptability for the weights pruning. The PMFS block offers a more lightweight solution than traditional self-attention mechanisms. Furthermore, its plug-and-play design facilitates easy integration into a variety of models without substantially adding to the computational overhead.

\subsection{Polarized Multi-scale Feature Self-attention (PMFS) Block}
Inspired by “polarized filtering” \cite{PSA}, We propose the Polarized Multi-scale Feature Self-attention (PMFS) block to tackle the limitation of traditional self-attention mechanisms that are computationally intensive especially when extended to 3D networks. The PMFS block is a plug-and-play block, which consists of three components in sequential order: Adaptive Multi-branch Feature Fusion (AMFF) layer, Polarized Multi-scale Channel Self-attention (PMCS) module, Polarized Multi-scale Spatial Self-attention (PMSS) module. The PMCS module enhances features in the channel dimension, while the PMSS module enhances features in the spatial dimension.

\subsubsection{Adaptive Multi-branch Feature Fusion (AMFF) Layer}

\begin{figure}
	\centering
	\includegraphics[width=0.99\linewidth]{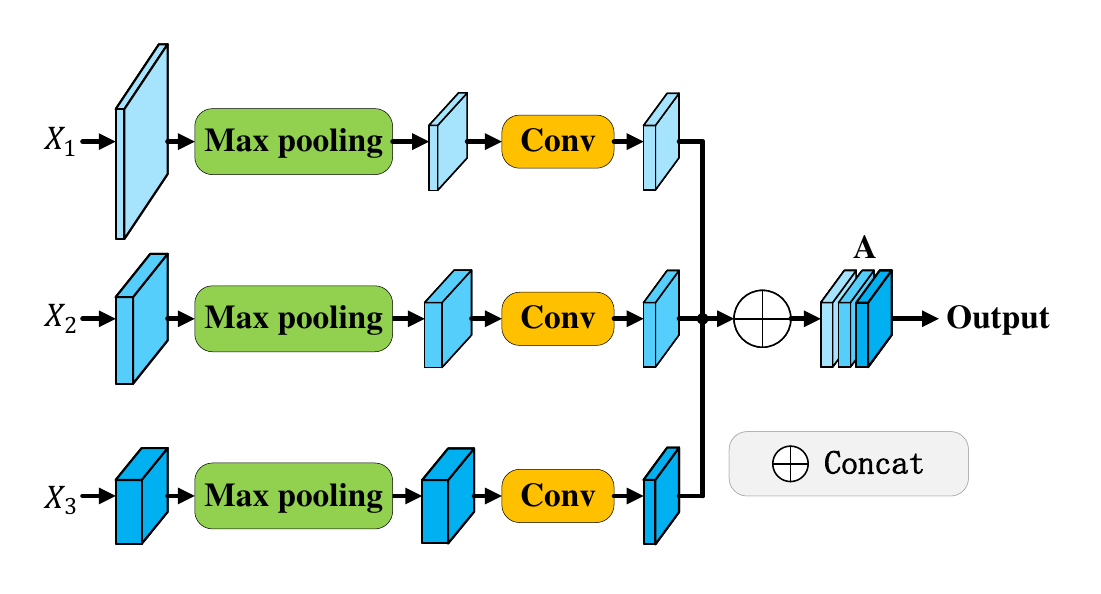}
	\caption{The Adaptive Multi-branch Feature Fusion (AMFF) layer. In one case, the resolutions of $X_1, X_2, X_3$ are $36\times80\times80\times48$, $64\times40\times40\times24$, $104\times20\times20\times12$. Downsampling and channel scaling unify resolution to $48\times20\times20\times12$, respectively. The resolution of fusion feature A is $144\times20\times20\times12$, which is obtained by concatenating the multi-branch features.}
    \label{fig: AMFF_Layer}
\end{figure}

Figure \ref{fig: AMFF_Layer} illustrates that, due to varying resolutions and channels at different stages, multi-branch features have first to be standardized to a uniform size using an adaptive multi-branch feature fusion layer, prior to performing feature enhancement.

The Adaptive Multi-branch Feature Fusion (AMFF) layer is capable of unifying diverse feature maps to the same channel and resolution. This is achieved by appropriately setting the channels, pooling kernel sizes, and the number of branches for the multi-branch features.

We denote the input features of different branches as $X_l\in\mathds{R}^{C_l \times H_l \times W_l \times D_l}(l\in\{1,2,3\})$, corresponding to the features of the $Stage_l(l\in\{1,2,3\})$ output. Max-pooling is first used to downsample $X_l$ to the same size as $X_3(X_3\in\mathds{R}^{C_3 \times H \times W \times D})$, and to unify the channels of the different branch features, the convolution is used to rescale the features. The specific operation is shown in the following equation($M_l\in\mathds{R}^{\frac{C}{3} \times H \times W \times D}$, $A\in\mathds{R}^{C \times H \times W \times D}$):

\begin{align}
    M_l&=W_l(K_l(X_l)), l\in\{1,2,3\} \\
    A&=M_1 \oplus M_2 \oplus M_3,
\end{align} where $K_1$, $K_2$, and $K_3$ are max-pooling with kernel sizes 4, 2, and 1, respectively. $W_1$, $W_2$, and $W_3$ refer to 3×3×3 convolution blocks. $\oplus$ denotes the concatenate operation.

\subsubsection{Polarized Multi-scale Channel Self-attention (PMCS) Module}

\begin{figure}
	\centering
	\includegraphics[width=0.99\linewidth]{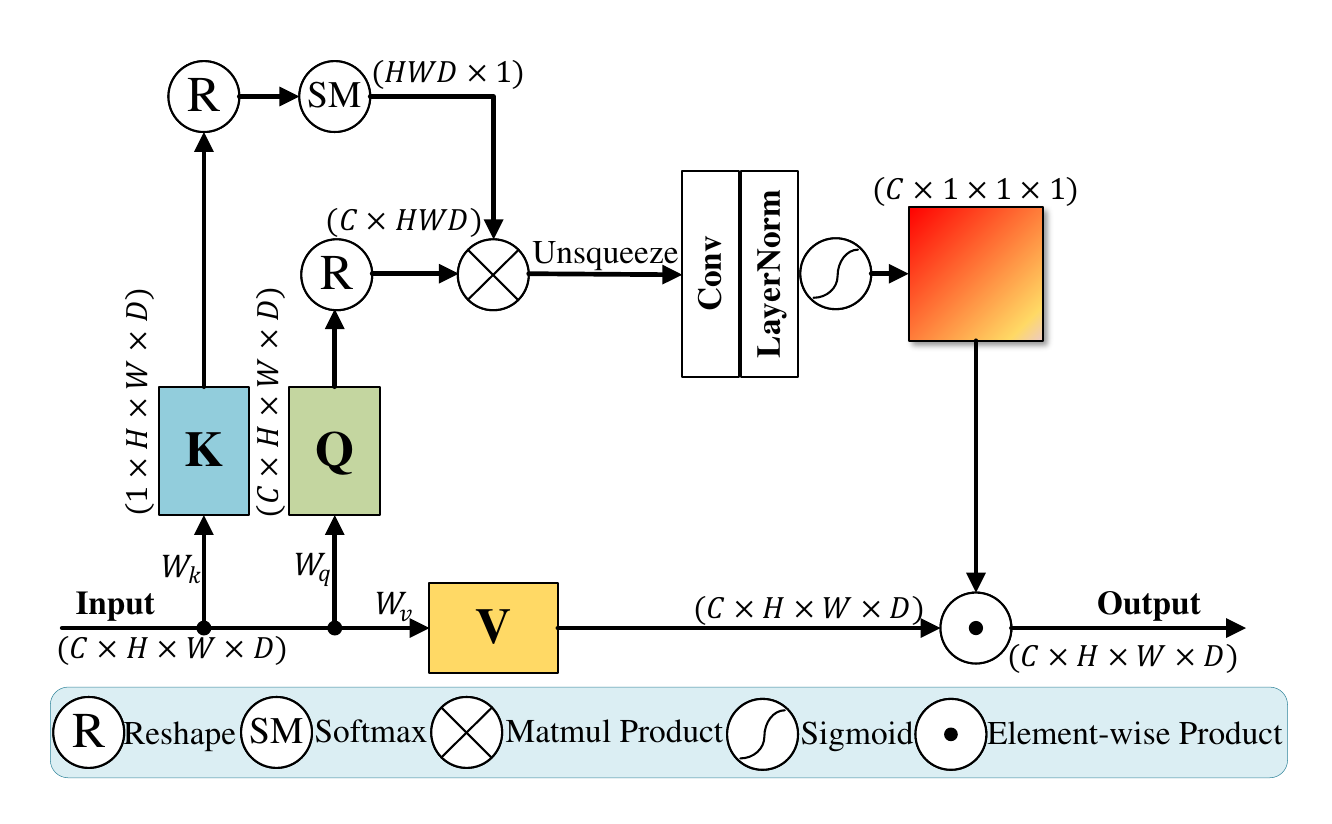}
	\caption{The Polarized Multi-scale Channel Self-attention (PMCS) module. In one case, the resolution of the input feature map is $C \times H \times W \times D (144 \times 20 \times 20 \times 12)$, where channel $C (48 + 48 + 48)$ is concatenated by three branches, whose channels are unified to 48. The depthwise separable convolution block is utilized to further decrease computational complexity.}
    \label{fig: PMCS_Module}
\end{figure}

As shown in Figure \ref{fig: PMCS_Module}, the PMCS module utilizes multi-scale feature information for expanding the number of attention points to capture global contextual features and more nuanced multi-scale channel dependencies. The PMCS module efficiently computes attention scores to enhance multi-scale channel features.

The channel of the input feature map is concatenated by three branches from different scales. The PMCS module combines branches and channels to expand the number of attention points (Figure \ref{fig: PMCS_Module}). Firstly, we compute each attention point's query and value vectors and the global channel key matrix $K^{ch}$. Then, the matrix $Q^{ch}$ consisting of the query vectors of all attention points and the matrix $K^{ch}$ are multiplied to obtain the multi-scale channel attention score matrix $Z^{ch}$. Finally, the matrix $Z^{ch}$ is multiplied element-wise by the matrix $V^{ch}$ to enhance the multi-scale channel features.

The input of the PMCS module is the fusion feature $A\in\mathds{R}^{C \times H \times W \times D}$, and the multi-scale channel attention score matrix $Z^{ch}\in\mathds{R}^{C \times 1 \times 1 \times 1}$ is defined as:

\begin{align}
    Q^{ch}&=\sigma_q\left(W_q\left(A\right)\right) \\
    K^{ch}&=F_{SM}\left(\sigma_k\left(W_k\left(A\right)\right)\right) \\
    Z^{ch}&=F_{SG}\left(F_{LN} \left(W_z\left(Q^{ch} \times K^{ch}\right)\right)\right),
\end{align} where $W_q$, $W_k$, $W_z$ are convolution layers, respectively. $\sigma_q$, $\sigma_k$ refer to different reshape operations. $F_{SM}(.)$ denotes the softmax operation $F_{\operatorname{SM}}(x)_i=\frac{e^{x_i}}{\sum_{j=1}^N e^{x_j}}$, $i\in{1,2,3,...N}$. $F_{LN}(.)$ represents layernorm operation. $F_{SG}(.)$ is the sigmoid activation function.

The output of the PMCS module is the feature map after multi-scale channel enhancement $A^{ch}\in\mathds{R}^{C \times H \times W \times D}$:

\begin{align}
    V^{ch}&=W_v\left(A\right) \\
    A^{ch}&=V^{ch} \odot Z^{ch},
\end{align} where $W_v$ is convolution layer, and $\odot$ denotes the element-wise product.

\subsubsection{Polarized Multi-scale Spatial Self-attention (PMSS) Module}

\begin{figure}
	\centering
	\includegraphics[width=0.99\linewidth]{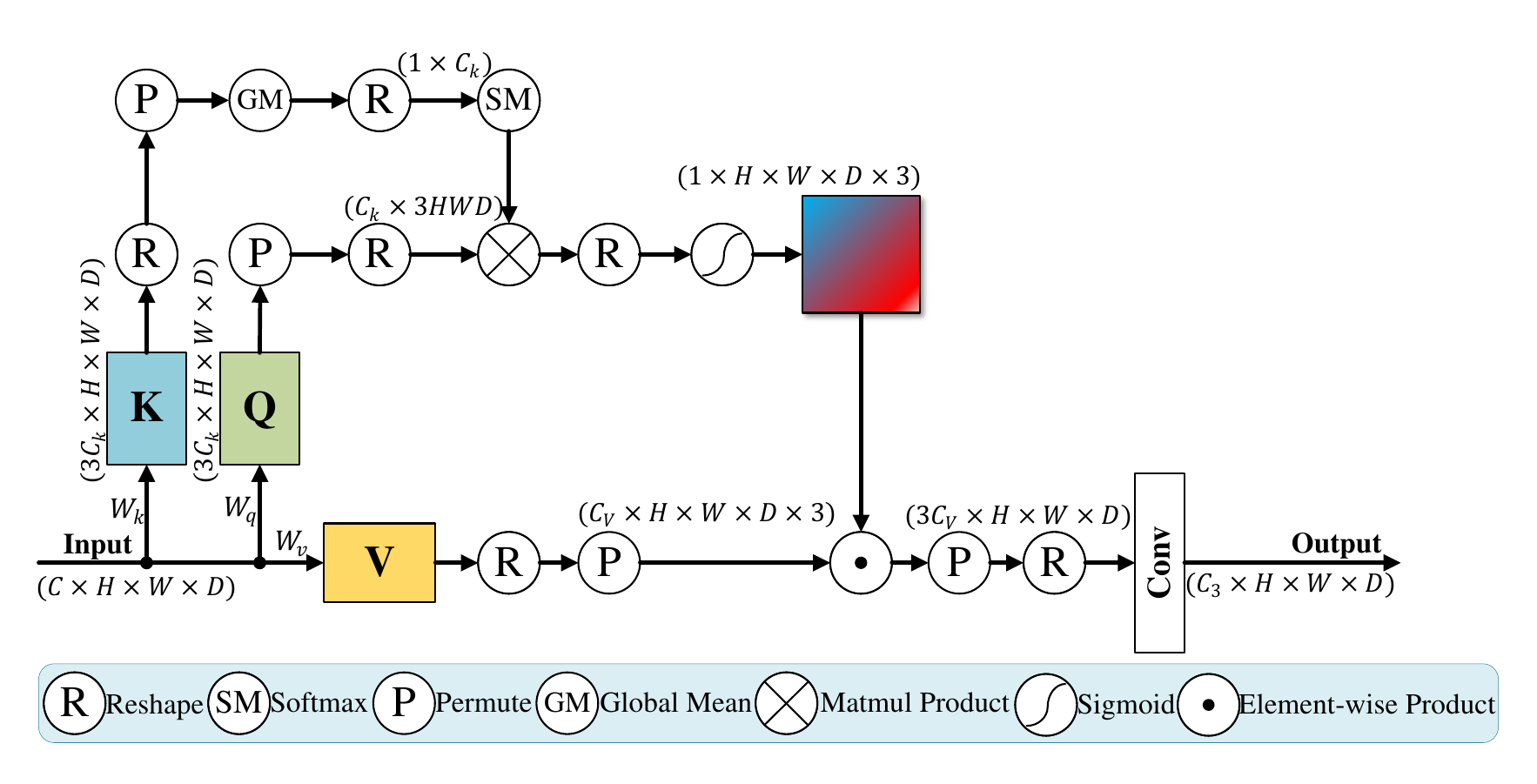}
	\caption{The Polarized Multi-scale Spatial Self-attention (PMSS) module. The PMSS module combines multi-branch spatial features, which means that it often needs to transform dimensions using the permute operation. It is worth noting that the bottleneck feature is eventually extracted from the enhanced multi-branch feature and has the same resolution as before the enhancement.}
    \label{fig: PMSS_Module}
\end{figure}

The PMSS module is designed to capture the more nuanced multi-scale spatial dependencies, which combines branches and spatial to expand the number of attention points (Figure \ref{fig: PMSS_Module}). Firstly, we compute each attention point's query and value vectors and the global spatial key matrix $K^{sp}$. Then, the matrix $K^{sp}$ and the matrix $Q^{sp}$ consisting of the query vectors of all attention points are multiplied to obtain the multi-scale spatial attention score matrix $Z^{sp}$. Finally, the matrix $Z^{sp}$ is multiplied element-wise by the matrix $V^{sp}$ to enhance the multi-scale spatial features.

At the end of the PMSS module, the bottleneck features are extracted from the enhanced multi-scale features. The advantage is that the enhancement of the features can be made more precise while obtaining a global context, resulting in a better semantic feature representation.

The input of the PMSS module is the feature map after multi-scale channel enhancement $A^{ch}\in\mathds{R}^{C \times H \times W \times D}$, multi-scale spatial attention score matrix $Z^{sp}\in\mathds{R}^{1 \times H \times W \times D \times 3}$:

\begin{align}
    Q^{sp}&=\sigma_q\left(P_q\left(W_q\left(A^{ch}\right)\right)\right) \\
    K^{sp}&=F_{SM}\left(\sigma_k^2\left(F_{G M}\left(P_k\left(\sigma_k^1\left(W_k\left(A^{ch}\right)\right)\right)\right)\right)\right) \\
    Z^{sp}&=F_{SG}\left(\sigma_{z}\left(K^{sp} \times Q^{sp}\right)\right),
\end{align} where $W_q$, and $W_k$ are convolution layers, respectively. $\sigma_q$, $\sigma_k^1$, $\sigma_k^2$, $\sigma_z$ refer to different reshape operations. $P_q$, $P_k$ denote dimension permute operations. $F_{GM}(.)$ represents global mean operation. $F_{SM}(.)$ denotes the softmax operation. $F_{SG}(.)$ is the sigmoid activation function.

The output of the PMSS module is the feature map after multi-scale spatial enhancement $A^{sp}\in\mathds{R}^{C_3 \times H \times W \times D}$ ($C_3 \times H \times W \times D$ corresponds to the resolution of the feature map $X_3$ output by $Stage_3$):

\begin{align}
    V^{sp}&=P_v\left(\sigma_v\left(W_v\left(A^{ch}\right)\right)\right) \\
    A^{sp}&=W_{\text{out}}\left(\sigma_{\text{out}}\left(\operatorname{P}_{\text{out}}\left(V^{sp} \odot Z^{sp}\right)\right)\right),
\end{align} where $W_v$, $W_{out}$ are convolution layers. $P_v$, $P_{out}$ refer to dimension permute operations. $\sigma_v$, $\sigma_{out}$ denote different reshape operations. $\odot$ is the element-wise product.

\subsection{Loss Function}
Class imbalance is a common issue in medical image segmentation, especially when random cropping is used during training. This approach may lead to the loss of representation for certain classes. To alleviate the above problem, we propose the weighted extended dice loss (WEDL). The WEDL can flexibly weight different classes and is smoother than the standard dice loss, which can be formulated as follows (The comparison with the standard dice loss is shown in Figure \ref{fig: loss}):

\begin{figure}
	\centering
	\includegraphics[width=0.99\linewidth]{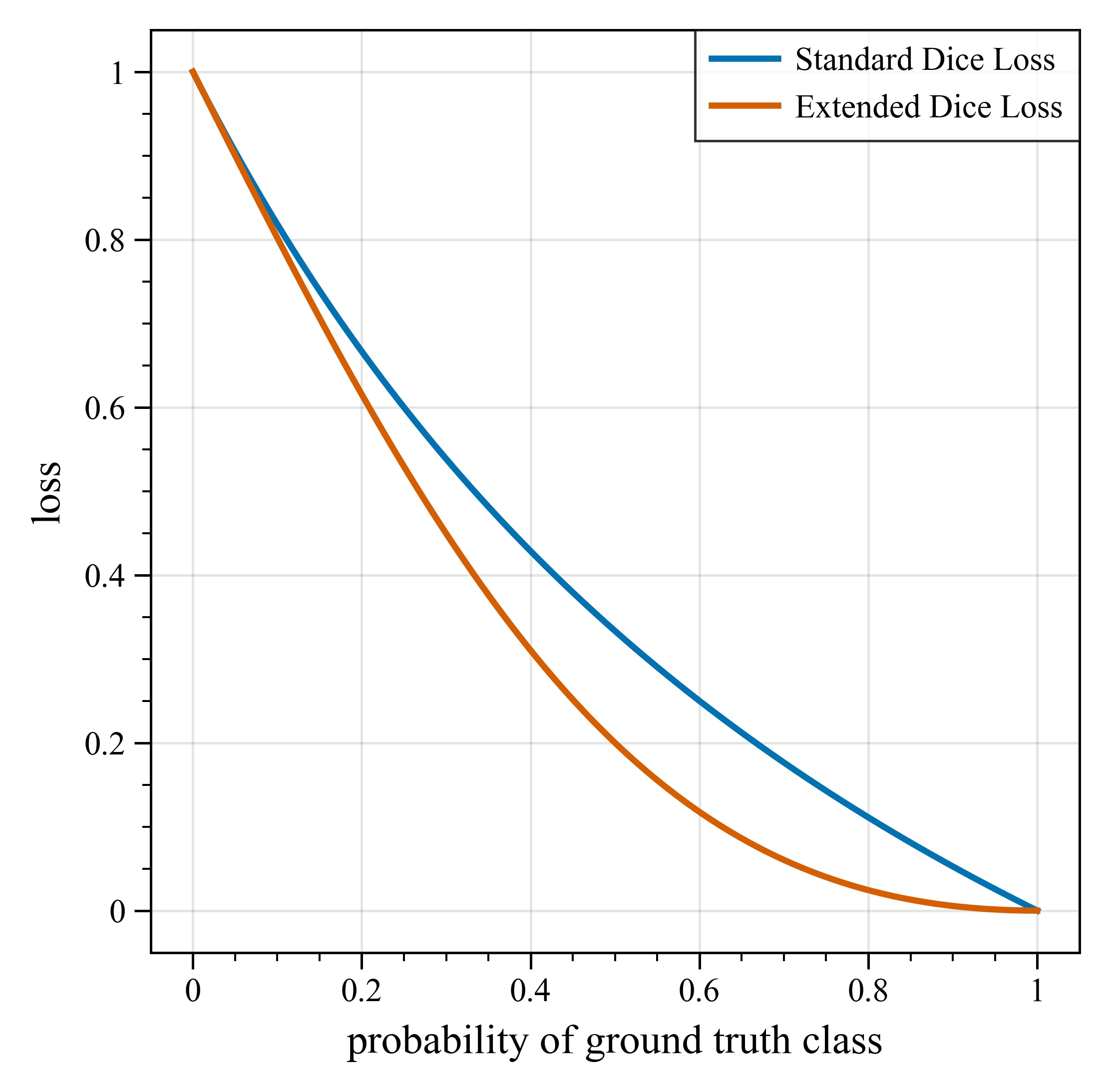}
	\caption{The WEDL vs. standard dice loss.}
    \label{fig: loss}
\end{figure}

\begin{equation}
    WEDL=1-\sum_{i=1}^{N_c} w_i \frac{2 \sum_{j=1}^{N_v} p_j g_j}{\sum_{j=1}^{N_v} p_j^2+\sum_{j=1}^{N_v} g_j^2+\varepsilon},
\end{equation} where the WEDL weights the $N_c$ classes and $w_i$ denotes the weight of the $\text{i-th}$ class. Each class sums over $N_v$ voxels, including the predicted segmentation volume $p_j\in P$ and the ground truth volume $g_j\in G$.

\section{Experiments}\label{Experiments}
Our method has been evaluated on three public datasets, encompassing multiple medical image modalities, including the tooth dataset from CBCT \cite{3D_CBCT_Tooth}, the MMOTU dataset from ultrasound images \cite{MMOTU}, and the ISIC 2018 dataset from dermoscopy images \cite{ISIC_2018_1, ISIC_2018_2}. We conduct extensive comparisons with a range of state-of-the-art (SOTA) methods, demonstrating the superiority of our proposed PMFSNet in achieving an optimal trade-off between accuracy and efficiency. Moreover, we also conducted an ablation study on the effectiveness and plug-and-play nature of the proposed PMFS block. 

\subsection{Dataset}
\subsubsection{3D CBCT Tooth}
The 3D CBCT tooth dataset is collected from a subset of a large-scale dataset \cite{3D_CBCT_Tooth}, which includes ground truth annotations. This dataset is used for segmentation and reconstruction of individual teeth and alveolar bone to aid in dental treatment (e.g., orthodontics, dental implants, and restoration). The large-scale dataset consists of large-scale CBCT imaging data from 15 different centers in China with varying data distributions. 129 scans are used in our experiments, which are divided into a training set of 103 scans and a test set of 26 scans. The data format of the original images is NIFTI, the voxel spacing is 0.25 mm, and the resolutions are $400\times400\times200$ voxels, $400\times400\times400$ voxels, and $320\times320\times320$ voxels. Some samples are in Figure \ref{fig: 3D_CBCT_Tooth_samples}.

\subsubsection{MMOTU}
The MMOTU dataset is a Multi-modality Ovarian Tumor Ultrasound (MMOTU) image dataset \cite{MMOTU}, which is commonly used for computer-aided diagnosing and ovarian tumor detection. It consists of two sub-sets with two modalities: OTU\_2d and OTU\_CEUS, respectively, including 1469 2D ultrasound images and 170 CEUS images. The MMOTU dataset is applied to different challenging tasks such as semantic segmentation and tumor recognition. We use 1000 images for training and 469 images for testing taken from task 1 of binary semantic segmentation. Particularly, the OTU\_2d sub-set contains 216 Color Doppler Flow Images (CDFI), where the rest of the 1253 images are traditional 2D ultrasound images, and the data format is all 2D RGB. In OTU\_2d, the width and height of images respectively range from 302\textasciitilde1135 pixels and 226\textasciitilde794 pixels. Some samples are in Figure \ref{fig: MMOTU_samples}.

\subsubsection{ISIC 2018}
The ISIC 2018 dataset is published by the International Skin Imaging Collaboration (ISIC) as a large-scale dataset of dermoscopy images in 2018 \cite{ISIC_2018_1, ISIC_2018_2}, which has become a major benchmark for the evaluation of medical image algorithms. The ISIC 2018 dataset is generally used for skin lesion analysis for melanoma detection. It contains 2594 dermoscopy images that are available at: https://challenge2018.isic-archive.com/. The data format of the ISIC 2018 dataset images is 2D RGB. We divide the 2594 images provided in task 1 of boundary segmentation into training (80\%) and test set (20\%), and we can see some samples in Figure \ref{fig: ISIC_2018_samples}.

\subsection{Evaluation Metrics}
We employ a variety of segmentation evaluation metrics to assess the network's performance from multiple perspectives, including Dice Similarity Coefficient (DSC), Intersection over Union (IoU), Mean Intersection over Union (mIoU), Accuracy (ACC), Hausdorff Distance (HD), Average Symmetric Surface Distance (ASSD), and Surface Overlap (SO).

\begin{equation}
    \begin{split}
        \operatorname{DSC}(P, G)&=\frac{2|P \cap G|}{|P|+|G|} \\
        &=\frac{2 \times T P}{(T P+F N)+(T P+F P)}
    \end{split}
\end{equation}
\begin{equation}
    IoU=\frac{TP}{TP+FN+FP}
\end{equation}
\begin{equation}
    mIoU=\frac{1}{n} \sum_{i=1}^n \frac{TP_i}{TP_i+FN_i+FP_i}
\end{equation}
\begin{equation}
    ACC=\frac{TP+TN}{TP+TN+FP+FN},
\end{equation} where TP, TN, FP, and FN are True Positive, True Negative, False Positive, and False Negative, respectively.

\begin{equation}
    \begin{split}
        d(p, S_G) = \min _{g \in S_G}\|p-g\|, p \in S_P
    \end{split}
\end{equation}
\begin{equation}
    \begin{split}
        HD(S_P, S_G)=\max \Big\{&\max _{p \in S_P} d(p, S_G), \\
        &\max _{g \in S_G} d(g, S_P)\Big\}
    \end{split}
\end{equation}
\begin{equation}
    \begin{split}
        \operatorname{ASSD}\left(S_P, S_G\right)=\frac{1}{|S_P|+|S_G|}&\left(\sum_{p \in S_P} d\left(p, S_G\right) \right. \\
        & \left. +\sum_{g \in S_G} d\left(g, S_P\right)\right)
    \end{split}
\end{equation}
\begin{subequations}
    \begin{equation}
        o(p)=\left\{\begin{array}{l}
        1, d\left(p, S_G\right)<\theta \\
        0, d\left(p, S_G\right)>\theta
        \end{array}\right.
    \end{equation}
    \begin{equation}
        SO\left(S_P\right)=\frac{\sum_{p \in S_P} o(p)}{\left|S_P\right|},
    \end{equation}
\end{subequations} where let $S_P$ be a set of surface voxels of predicted volumes, and $S_G$ be a set of surface voxels of ground truth volumes. $\|.\|$ is the distance paradigm between point sets, e.g. Euclidean distance. $|.|$ is the number of points of the set. $\theta$ is a maximal distance to determine whether two points have the same spatial position.

\subsection{Implementation Details}
All experiments are conducted using Pytorch (version 1.12) and on a Ubuntu 18.04 LTS workstation operation system with a 3.70GHz i7-8700K CPU and a 32G V100 GPU.

For the 3D CBCT tooth dataset, we resample each 3D image to a uniform voxel spacing of $0.5 \times 0.5 \times 0.5$mm and then randomly crop every image to $160 \times 160 \times 96$ size. We also adopt some data augmentation methods such as random elastic deformation, adding Gaussian noise, random flipping, random scaling, random rotation, and random shift. Because of the large range of image Hounsfiled Unit (HU) values, we clip the lower and upper bounds of the image values to the range -1412 to 17943. All images are standardized and normalized to fit within an intensity range of [0, 1]. To improve the stability of training, we employ the trick of gradient accumulation. We employ the Adam optimizer, which is initialized with a learning rate of 0.0005 and a weight decay of 0.00005. The learning rate is tuned during training using the ReduceLROnPlateau strategy. All networks are trained for 20 epochs with a batch size of 1.

For the 2D images, we resize each image to $224 \times 224$ pixels and use data augmentation methods such as random resize and crop, color jitter, random Gaussian noise, random flipping, random rotation, and cutout \cite{Cutout}. The batch size is 32 and we employ the AdamW optimizer. On the MMOTU dataset, All networks are trained for a total of 2000 epochs with an initial learning rate of 0.01 and weight decay of 0.00001, and the CosineAnnealingLR strategy is used to tune the learning rate. To ensure comparability with other methods, the PMFSNet is pre-trained on the 1000-class ImageNet 2012 dataset \cite{ImageNet2012} by default. On the ISIC 2018 dataset, All networks are trained for a total of 150 epochs with an initial learning rate of 0.005 and weight decay of 0.000001, and the CosineAnnealingWarmRestarts strategy is used to tune the learning rate.

We propose several scaling versions of the model structure for tasks with various data sizes, and the corresponding scaling versions can be specified by modifying the hyperparameters. They are differentiated by the number of dense-feature-stacking units per stage, the channels per stage, and whether they have a decoder (Figure \ref{fig: PMFSNet}). According to the parameters of these scaling versions in descending order, they are named BASIC, SMALL, and TINY. In addition, the models are referred to as PMFSNet2D and PMFSNet3D according to whether the input dimensions are 2D or 3D. The parameter count, computational complexity, and experimental performance between the different scaling versions are described in detail in Appendix A.

\subsection{Quantitative Evaluations}
We conduct comparative experiments with various state-of-the-art (SOTA) networks on each of the three public datasets to quantify the superiority of our proposed network in terms of efficiency and performance. Our method is adaptable to both 2D and 3D medical images demonstrating the scalability and portability of our proposed network.

\subsubsection{Comparisons with the state-of-the-arts on 3D CBCT tooth}

\begin{table*}[H]
\centering
\caption{Comparison results of different methods on the 3D CBCT tooth dataset. The best results are in bold. $\uparrow$ means higher values are better, $\downarrow$ means lower values are better.}
\label{tab: 3D CBCT Tooth Comparison}
\setlength{\tabcolsep}{2.2mm}{
    \begin{tabular}{lccccccc}
    \toprule
    Method & FLOPs(G)$\downarrow$ & Params(M)$\downarrow$ & HD(mm)$\downarrow$ & ASSD(mm)$\downarrow$ & IoU(\%)$\uparrow$ & SO(\%)$\uparrow$ & DSC(\%)$\uparrow$ \\
    \midrule
    UNet3D \cite{UNet3D} & 2223.03 & 16.32 & 113.79 & 22.40 & 70.62 & 70.72 & 36.67 \\
    DenseVNet \cite{DenseVNet} & 23.73 & 0.87 & 8.21 & 1.14 & 84.57 & 94.88 & 91.15 \\
    AttentionUNet3D \cite{AttentionUNet} & 2720.79 & 94.48 & 147.10 & 61.10 & 52.52 & 42.49 & 64.08 \\
    DenseVoxelNet \cite{DenseVoxelNet} & 402.32 & 1.78 & 41.18 & 3.88 & 81.51 & 92.50 & 89.58 \\
    MultiResUNet3D \cite{MultiResUNet3D} & 1505.38 & 17.93 & 74.06 & 8.17 & 76.19 & 81.70 & 65.45 \\
    UNETR \cite{UNETR} & 229.19 & 93.08 & 107.89 & 17.95 & 74.30 & 73.14 & 81.84 \\
    SwinUNETR \cite{SwinUNETR} & 912.35 & 62.19 & 82.71 & 7.50 & 83.10 & 86.80 & 89.74 \\
    TransBTS \cite{TransBTS} & 306.80 & 33.15 & 29.03 & 4.10 & 82.94 & 90.68 & 39.32 \\
    nnFormer \cite{nnFormer} & 583.49 & 149.25 & 51.28 & 5.08 & 83.54 & 90.89 & 90.66 \\
    3D UX-Net \cite{3DUXNet} & 1754.79 & 53.01 & 108.52 & 19.69 & 75.40 & 73.48 & 84.89 \\
    \midrule
    \textbf{PMFSNet3D (Ours)} & \textbf{15.14} & \textbf{0.63} & \textbf{5.57} & \textbf{0.79} & \textbf{84.68} & \textbf{95.10} & \textbf{91.30} \\
    \bottomrule
    \end{tabular}}
\end{table*}

\begin{figure*}
	\centering
	\includegraphics[width=0.99\linewidth]{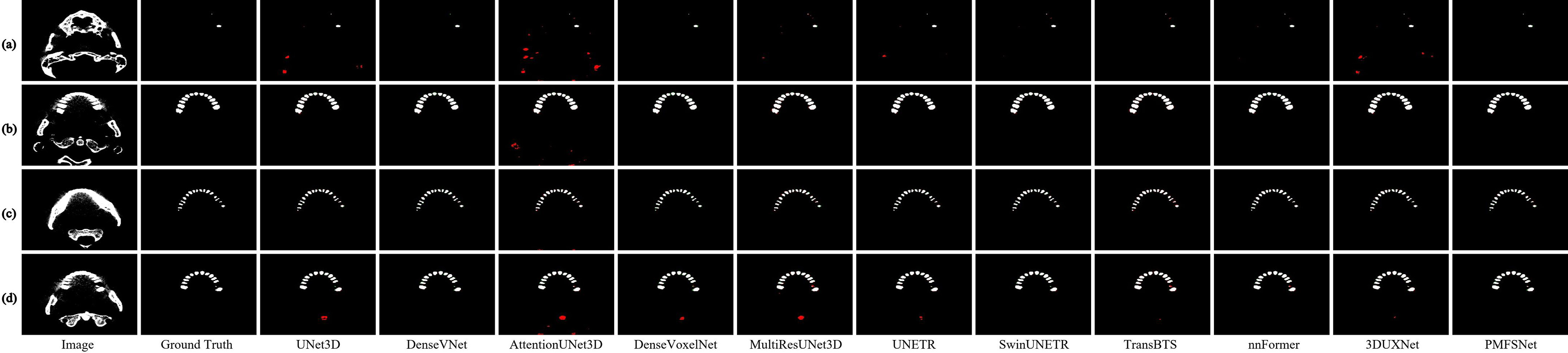}
	\caption{Visual comparison with the state-of-the-arts on the 3D CBCT tooth dataset. The colors white, green, and red represent the correct segmentation, the under-segmentation, and the over-segmentation, respectively.}
    \label{fig: 3D_CBCT_Tooth_segmentation}
\end{figure*}

To evaluate the performance of the PMFSNet on a 3D CBCT tooth dataset, we comprehensively compare it with ten state-of-the-art (SOTA) 3D networks. These networks include UNet3D \cite{UNet3D} based on traditional UNet architectures, MultiResUNet3D \cite{MultiResUNet3D} based on residual structures, DenseVNet \cite{DenseVNet} and DenseVoxelNet \cite{DenseVoxelNet} based on dense feature stacking, AttentionUNet3D \cite{AttentionUNet} based on attention mechanism, and transformer-based UNETR \cite{UNETR}, SwinUNETR \cite{SwinUNETR}, TransBTS \cite{TransBTS}, nnFormer \cite{nnFormer}, 3D UX-Net \cite{3DUXNet}. The comparison results are shown in Table \ref{tab: 3D CBCT Tooth Comparison}. 

The results presented in Table \ref{tab: 3D CBCT Tooth Comparison} demonstrate the superior efficiency and effectiveness of PMFSNet in 3D CBCT segmentation tasks compared to other methods (Params-FLOPs-IoU correlation comparison in Figure \ref{fig: 3D_CBCT_Tooth_bubble_image}). It achieves the best results in all metrics compared to current state-of-the-art methods. Notably, the PMFSNet outperforms the SOTA method DenseVNet in all aspects. More specifically, PMFSNet achieves a reduction in Floating-point Operations Per Second (FLOPs) and Parameters (Params) by 8.59 GFLOPs and 0.24 M, respectively. Additionally, it shows improvements in various metrics: it reduces the Hausdorff Distance (HD) by 2.64 mm and the Average Symmetric Surface Distance (ASSD) by 0.35 mm. Moreover, PMFSNet enhances the Intersection over Union (IoU) by 0.11\%, the Similarity Overlap (SO) by 0.22\%, and the Dice Similarity Coefficient (DSC) by 0.15\%.

The qualitative study in Figure \ref{fig: 3D_CBCT_Tooth_segmentation} illustrates the comparison of the performance of different networks on the 3D CBCT tooth dataset. The images in Figure \ref{fig: 3D_CBCT_Tooth_segmentation}(a) show the segmentation results of different networks for the cusps of the teeth. We can observe that our method can accurately segment tiny regions, while the comparison methods often lead to over-segmentation. The images in Figure \ref{fig: 3D_CBCT_Tooth_segmentation}(c) present the segmentation results of different networks for uneven surfaces of crowns. Compared to other networks, PMFSNet retains more detail for irregularly shaped surface contours. The images in Figure \ref{fig: 3D_CBCT_Tooth_segmentation}(d) present the segmentation results of different networks for cases where there are missing teeth and being affected by other bone tissue. In this challenging situation, PMFSNet demonstrates promising segmentation performance and does not over-segment other bone tissue or areas of missing teeth.

\subsubsection{Comparisons with the state-of-the-arts on MMOTU}

\begin{table*}[H]
\centering
\caption{Comparison results of different methods on the MMOTU dataset. The best results are in bold. $\uparrow$ means higher values are better, $\downarrow$ means lower values are better.}
\label{tab: MMOTU Comparison}
\setlength{\tabcolsep}{5.3mm}{
    \begin{tabular}{lccccc}
    \toprule
    Method & FLOPs(G)$\downarrow$ & Params(M)$\downarrow$ & IoU(\%)$\uparrow$ & mIoU(\%)$\uparrow$ & Iterations \\
    \midrule
    PSPNet \cite{PSPNet} & 38.71 & 53.32 & 82.01 & 89.41 & 20k \\
    DANet \cite{DANet} & 10.95 & 47.44 & 82.20 & 89.53 & 20k \\
    SegFormer \cite{SegFormer} & 2.52 & 7.72 & \textbf{82.46} & \textbf{89.88} & 80k \\
    U-Net \cite{UNet} & 41.93 & 31.04 & 79.91 & 86.80 & 80k \\
    TransUNet \cite{TransUNet} & 24.61 & 105.28 & 81.31 & 89.01 & 80k \\
    BiseNetV2 \cite{BiseNetV2} & 3.40 & 5.19 & 79.37 & 86.13 & 80k \\
    \midrule
    \textbf{PMFSNet2D (Ours)}& \textbf{2.21} & \textbf{0.99} & 82.02 & 89.36 & 2k \\
    \bottomrule
    \end{tabular}}
\end{table*}

\begin{figure}
	\centering
	\includegraphics[width=0.99\linewidth]{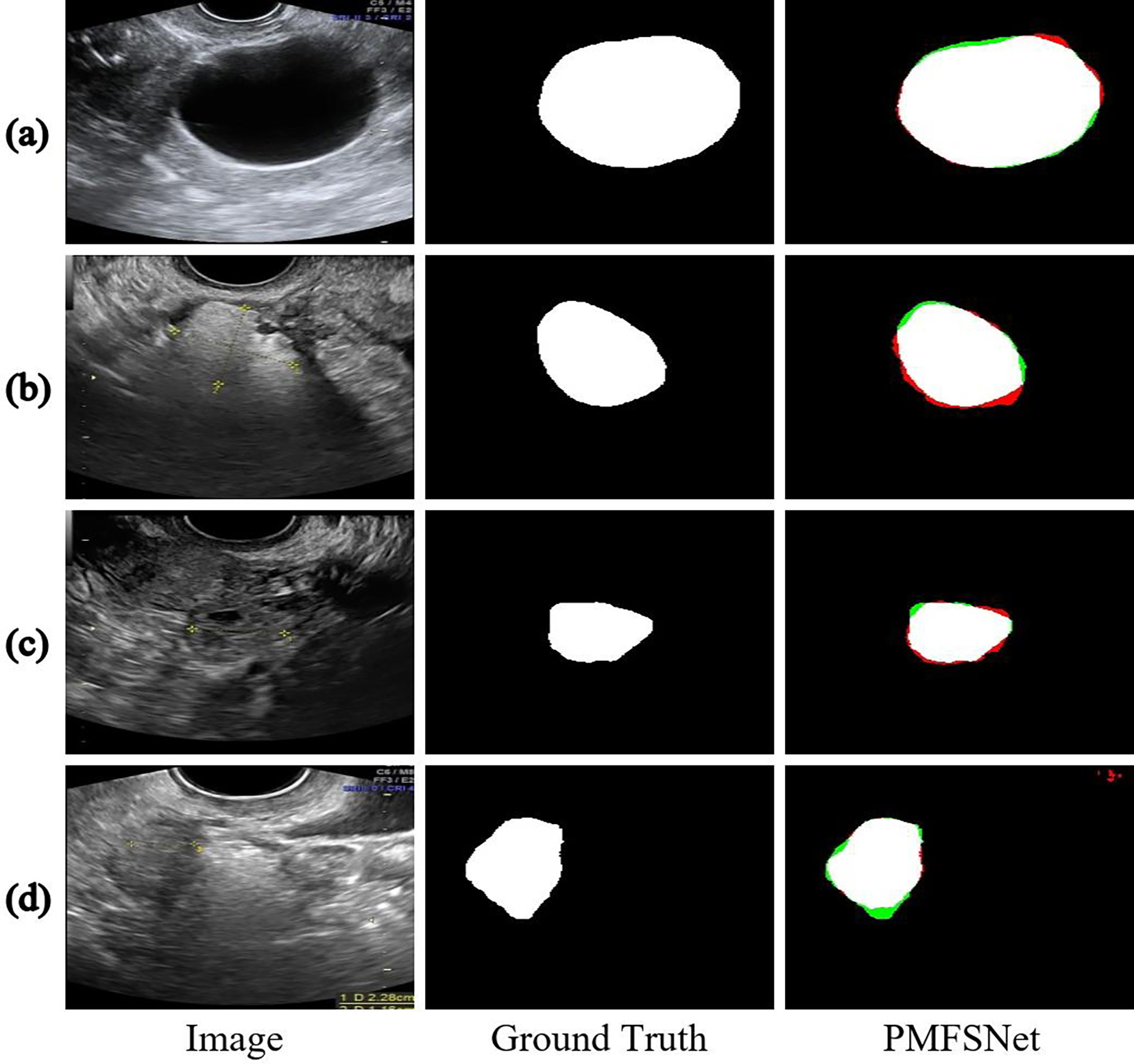}
	\caption{Visual segmentation performance of the PMFSNet on the MMOTU dataset. The colors white, green, and red represent the correct segmentation, the under-segmentation, and the over-segmentation, respectively.}
    \label{fig: MMOTU_segmentation}
\end{figure}

On the MMOTU dataset, our proposed PMFSNet is evaluated against six networks used in study \cite{MMOTU}, including PSPNet \cite{PSPNet}, DANet \cite{DANet}, SegFormer \cite{SegFormer}, U-Net \cite{UNet}, TransUNet \cite{TransUNet}, and BiseNetV2
\cite{BiseNetV2}. To ensure a fair comparison in our study with the MMOTU dataset, we pre-train the PMFSNet on the 1000-class ImageNet 2012 dataset \cite{ImageNet2012}. This approach mirrors the methodology used in the cited experiment, where each network is also loaded with pre-training weights.

We can observe from Table \ref{tab: MMOTU Comparison}, in terms of segmentation performance, SegFormer (with 7.72 M parameters) leads with the highest scores in IoU and mIoU metrics, achieving 82.46\% and 89.88\%, respectively. PMFSNet closely follows, attaining an IoU of 82.02\% and a mIoU of 89.36\%, closely competing with the highest-performing networks in segmentation with significantly less parameter count. More specifically, the PMFSNet records the lowest (FLOPs) at 2.21 GFLOPs and maintains the smallest model size with only 0.99 million parameters. In comparison, BiseNetV2 has a relatively small parameter count, yet it is still 5 times larger than PMFSNet. SegFormer, despite its superior segmentation performance, has a model size more than 7 times larger than PMFSNet, highlighting PMFSNet's advantage in terms of efficiency and compactness.

Figure \ref{fig: MMOTU_segmentation} illustrates the quantitative results of the PMFSNet on the MMOTU dataset. We can observe that our PMFSNet can achieve promising segmentation performance even in various challenging cases. The images in Figure \ref{fig: MMOTU_segmentation}(a) show the segmentation result of a general sample, where PMFSNet can accurately segment the ovarian lesion region. The images in Figure \ref{fig: MMOTU_segmentation}(b) show the segmentation result of a sample with an inconspicuous lesion region. It is extremely challenging to identify the lesion region from the original image accurately. The images in Figure \ref{fig: MMOTU_segmentation}(c) show the segmentation result of a lesion region with blurred boundaries, where the original image has many dark regions of interference. The images in Figure \ref{fig: MMOTU_segmentation}(d) show the segmentation results of a low-contrast sample in which the lesion region is not clearly distinguishable.

\subsubsection{Comparisons with the state-of-the-arts on ISIC 2018}

\begin{table*}[H]
\centering
\caption{Comparison results of different methods on the ISIC 2018 dataset. The best results are in bold. $\uparrow$ means higher values are better, $\downarrow$ means lower values are better.}
\label{tab: ISIC 2018 Comparison}
\setlength{\tabcolsep}{5.3mm}{
    \begin{tabular}{lccccc}
    \toprule
    Method & FLOPs(G)$\downarrow$ & Params(M)$\downarrow$ & IoU(\%)$\uparrow$ & DSC(\%)$\uparrow$ & ACC(\%)$\uparrow$ \\
    \midrule
    U-Net \cite{UNet} & 41.93 & 31.04 & 76.77 & 86.55 & 95.00 \\
    AttU-Net \cite{AttentionUNet} & 51.07 & 34.88 & 78.19 & 87.54 & 95.33 \\
    CA-Net \cite{CANet}  & 4.62 & 2.79 & 68.82 & 80.96 & 92.96 \\
    BCDU-Net \cite{BCDUNet}  & 31.96 & 18.45 & 76.46 & 86.26 & 95.19 \\
    CE-Net \cite{CENet} & 6.83 & 29.00 & 78.05 & 87.47 & 95.40 \\
    CPF-Net \cite{CPFNet} & 6.18 & 43.27 & 78.47 & 87.70 & 95.52 \\
    CKDNet \cite{CKDNet} & 12.69 & 59.34 & 77.89 & 87.35 & 95.27 \\
    \midrule
    \textbf{PMFSNet2D (Ours)}& \textbf{2.21} & \textbf{0.99} & \textbf{78.82}  & \textbf{87.92} & \textbf{95.59} \\
    \bottomrule
    \end{tabular}}
\end{table*}

\begin{figure*}
	\centering
	\includegraphics[width=0.99\linewidth]{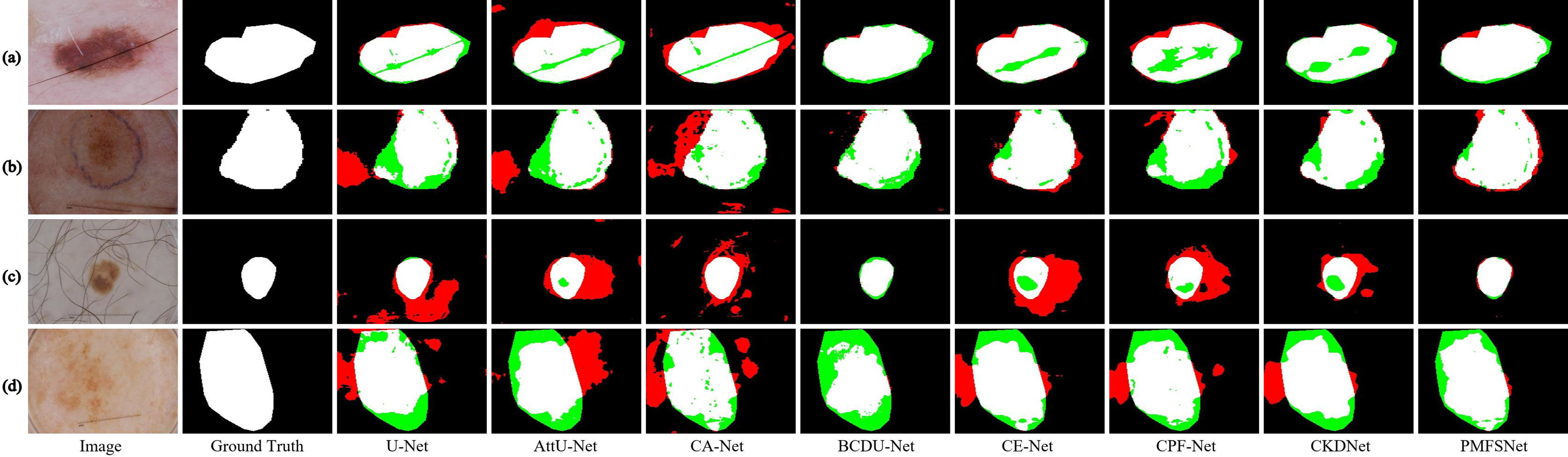}
	\caption{Visual comparison with the state-of-the-arts on the ISIC 2018 dataset. The colors white, green, and red represent the correct segmentation, the under-segmentation, and the over-segmentation, respectively.}
    \label{fig: ISIC2018_segmentation}
\end{figure*}

We ensure the consistency of the experimental setup on the ISIC 2018 dataset and implement a fair comparison experiment. The networks compared with our PMFSNet include two general segmentation networks and five networks specifically designed for skin lesions. We re-implement U-Net \cite{UNet}, AttU-Net \cite{AttentionUNet}, CA-Net \cite{CANet}, BCDU-Net \cite{BCDUNet}, CE-Net \cite{CENet}, CPF-Net \cite{CPFNet}, and CKDNet \cite{CKDNet}.

The results presented in Table \ref{tab: ISIC 2018 Comparison} demonstrate the superior efficiency and accuracy of PMFSNet in medical image segmentation tasks compared to other methods. Notably, the PMFSNet outperforms the SOTA method AttU-Net in generalized segmentation networks and the SOTA method CPF-Net in specialized segmentation networks in all aspects. More specifically, PMFSNet achieves a reduction in Floating-point Operations Per Second (FLOPs) against AttU-Net and CPF-Net by 48.86 GFLOPs and 3.97 GFLOPs respectively, and a reduction in Parameters (Params) by 33.89 M and 42.28 M respectively. Additionally, it shows improvements in other metrics: it enhances the Intersection over Union (IoU) by 0.63\% and 0.35\% respectively, the Dice Similarity Coefficient (DSC) by 0.38\% and 0.22\% respectively, and the Accuracy (ACC) by 0.26\% and 0.07\% respectively. These results demonstrate that our PMFSNet as a general segmentation network is more generalized and can outperform dedicated networks in the skin lesion segmentation task. In addition, the result also demonstrates the feasibility of obtaining better segmentation performance with a much smaller model.

Qualitative analysis in Figure \ref{fig: ISIC2018_segmentation} illustrates the comparison of the performance of our method to different networks on the ISIC 2018 dataset. These samples in Figure \ref{fig: ISIC2018_segmentation} contain various challenges, either occluded lesions, low-light conditions, low-contrast lesions, or blurred lesion boundaries. More specifically, the images in Figure \ref{fig: ISIC2018_segmentation}(a) show the segmentation results of different networks for occluded lesions. It can be seen that the comparison methods often lead to over-segmentation when it comes to occluded lesions. The images in Figure \ref{fig: ISIC2018_segmentation}(b) display the segmentation results of different networks for lesions under a low-light condition. Credits to the PMFS block's capability to enhance crucial features at the network's bottleneck, the PMFSNet delivers superior segmentation performance, particularly for lesions under a low-light condition, with a notable improvement in edge delineation. The images in Figure \ref{fig: ISIC2018_segmentation}(c) present the segmentation results of different networks for lesions with artifacts. Compared to other networks, our PMFSNet effectively avoids segmentation errors that could be caused by the interference of hairs in the images. The images in Figure \ref{fig: ISIC2018_segmentation}(d) present the segmentation results of different networks for lesions with blurred boundaries and low-contrast lesions. Even in such challenging conditions, PMFSNet maintains strong generalization performance.

\subsection{Ablation Studies}

\begin{table}[H]
\centering
\caption{The effect of PMFS block on segmentation performance on our models. The best results are in bold. $\uparrow$ means higher values are better, $\downarrow$ means lower values are better. \checkmark means with PMFS block, $\times$ means without PMFS block.}
\label{tab: PMFS block our ablation}
\setlength{\tabcolsep}{1.8mm}{
    \begin{tabular}{cccc}
    \toprule
    Dataset & PMFS block & Params(M)$\downarrow$ & IoU(\%)$\uparrow$ \\
    \midrule
    \multirow{2}{*}{3D CBCT tooth} & $\times$ & \textbf{0.37} & 82.00 \\
       & \checkmark & 0.63 & \textbf{84.68} \\
    \midrule
    \multirow{2}{*}{MMOTU} & $\times$ & \textbf{0.65} & 80.45 \\
       & \checkmark & 0.99 & \textbf{82.02} \\
    \midrule
    \multirow{2}{*}{ISIC 2018} & $\times$ & \textbf{0.65} & 77.56 \\
       & \checkmark & 0.99 & \textbf{78.82} \\
    \bottomrule
    \end{tabular}}
\end{table}

To verify the effectiveness of the PMFS block in enhancing our PMFSNet model's segmentation capabilities, we carry out ablation experiments on various datasets. The results of the ablation experiments on our proposed model regarding PMFS blocks are shown in Table \ref{tab: PMFS block our ablation}. By embedding the PMFS block into two benchmark architectures, PMFSNet3D and PMFSNet2D, we observe an increase in Parameters (Params) of 0.26 M and 0.34 M, respectively. Despite a minor increase in the Parameters (Params), the resulting performance improvements are significant. Moreover, on the three datasets, the PMFSNet model with PMFS block exhibits significant performance gains, with IoU improvements of 2.68\%, 1.57\%, and 1.26\%, respectively. These results verify the effectiveness of the PMFS block in our proposed method.

To further illustrate that the PMFS block is both plug-and-play and remains effective when integrated with other models, we conduct ablation experiments applying the PMFS block on various other networks. We select the three models, i.e., U-Net, CA-Net, and BCDU-Net, in Table \ref{tab: ISIC 2018 Comparison}. Subsequently, we perform a quantitative analysis of the segmentation performance for these three models, comparing their results before and after the integration of the PMFS block.

\begin{table}[H]
\centering
\caption{The effect of PMFS block on segmentation performance on other models. The best results are in bold. $\uparrow$ means higher values are better, $\downarrow$ means lower values are better. \checkmark means with PMFS block, $\times$ means without PMFS block.}
\label{tab: PMFS block other ablation}
\setlength{\tabcolsep}{0.8mm}{
    \begin{tabular}{lcccc}
    \toprule
    Method & PMFS block & FLOPs(G)$\downarrow$ & Params(M)$\downarrow$ & IoU(\%)$\uparrow$ \\
    \midrule
    \multirow{2}{*}{U-Net} & $\times$ & 41.93 & \textbf{31.04} & 76.77 \\
       & \checkmark & 42.16 & 32.25 & \textbf{77.21} \\
    \midrule
    \multirow{2}{*}{CA-Net} & $\times$ & 4.62 & \textbf{2.79} & 68.82 \\
       & \checkmark & 4.77 & 3.64 & \textbf{74.48} \\
    \midrule
    \multirow{2}{*}{BCDU-Net} & $\times$ & 31.96 & \textbf{18.45} & 76.46 \\
       & \checkmark & 32.44 & 19.12 & \textbf{76.87} \\
    \bottomrule
    \end{tabular}}
\end{table}

Table \ref{tab: PMFS block other ablation} demonstrates that integrating PMFS blocks into U-Net, CA-Net, and BCDU-Net results in only a marginal increase in Floating-point Operations Per Second (FLOPs) and Parameters (Params). This indicates that the addition of the PMFS block does not significantly impact the computational efficiency of these models. Furthermore, the results reveal that the PMFS block leads to an improvement in the Intersection over Union (IoU) metric for U-Net, CA-Net, and BCDU-Net models, achieving 77.21\%, 74.48\%, and 76.87\%, respectively. These experimental results suggest that the PMFS block, when integrated into other models, can effectively bolster their feature representation capabilities.

\section{Discussion}\label{Discussion}
Current pure Vision Transformers (ViTs) in medical image segmentation have shown the ability to learn long-term dependencies, but this often comes at the expense of increased model complexity and a loss of the inductive bias inherent in CNNs. While some studies have explored hybridizing transformers and CNNs to model both long-term and local dependencies \cite{ContNet}, the performance of these large hybrid models can be limited by the typically smaller size of medical image datasets compared to natural image datasets. Large models, whether based on transformers, deep CNNs, or a hybrid, tend to be data-hungry. In contrast, medical datasets are relatively smaller, leading to a higher risk of overfitting during training. Table \ref{tab: 3D CBCT Tooth Comparison} presents various models based on transformer architecture along with their evaluation results, highlighting this issue.

Given this context, we posit that developing a lightweight network for medical image segmentation is crucial and imperative to tackle the problem of redundant computation found in larger models. Our comprehensive results demonstrate that even with a model ($\leq$ 1 million parameters), superior performance can be attained in various segmentation tasks across different data scales.

In this work, we optimize the network architecture by employing a streamlined 3-stage encoder and introducing self-attention computation exclusively at the network's bottleneck. We propose a plug-and-play PMFS block to encode long-term dependencies for feature enhancement. To enhance efficiency, we optimize the computational complexity of the attention score matrix through "polarized filtering" and replace standard convolution with depthwise separable convolution. Lastly, we integrate multi-scale features into global channel and spatial dimensions, thereby increasing the number of attention points and enhancing the feature representation.

The discourse on encoding local and long-term dependencies is intrinsically linked to the comparative studies between Convolutional Neural Networks (CNNs) and transformers, as well as the exploration of hybrid mechanisms combining these two approaches \cite{ConvNets, ConvMixer}. CNNs excel in encoding local dependencies due to their inherent architectural design, which emphasizes local receptive fields and hierarchical feature extraction. In contrast, transformers, with their self-attention mechanisms, are adept at capturing long-term dependencies by considering global interactions across an entire input sequence. The hybrid of these two architectures aims to leverage the strengths of both CNNs and ViTs: the local feature extraction prowess of CNNs and the global contextual awareness of transformers.

Based upon these pioneer works along with our findings reveal that while CNNs learn hierarchical features in a sequential layer-by-layer manner, each layer is processed at a different level of abstraction, implying that shallow features are extracted without direct guidance from deeper layers in the network. In our method, we are trying to mutually enhance multi-scale features with the idea of parallel computing in the self-attention mechanism. This is due to the fact that transformers employ a parallel computing approach in their self-attention mechanism, focusing on simultaneous feature learning across both channel and spatial dimensions, but, normally only at the same scale. In this case, our approach seeks to harness this parallelism for multi-scale feature learning with greater granularity, diverging from the inherent sequential process of CNNs to achieve a more comprehensive and nuanced understanding of feature interactions.

In terms of model portability, the proposed lightweight model is conducive to deployment on edge devices and enhances the data stream in current medical imaging equipment, especially during the inference phase. This feature is crucial for real-time applications and plays a significant role in improving the efficiency and accessibility of medical imaging technology across diverse environments. In the future, we will further evaluate the model performance on mobile and edge devices and extend the model to more specific instance segmentation tasks.

\section{Conclusion}\label{Conclusion}
In this paper, we propose a novel lightweight semantic segmentation network (PMFSNet) designed to adapt to a variety of medical image segmentation tasks in different image modalities and take into account the balance between efficiency and performance. We optimize the network architecture by employing a streamlined 3-stage encoder and introducing self-attention computation exclusively at the network's bottleneck. The proposed plug-and-play PMFS block is a multi-scale feature enhancement module based on the self-attention mechanism, to encode long-term dependencies and incorporate global contextual features. To enhance efficiency, we optimize the computational complexity of the attention score matrix and adopt the depthwise separable convolution. The experiments on three public datasets containing both 2D and 3D modalities show that PMFSNet has strong application scalability for a variety of medical image segmentation tasks. The proposed method achieves competitive segmentation performance compared to the current SOTA methods across several 2D and 3D medical imaging segmentation tasks, achieving this with 27.6\% fewer parameters than DenseVNet, 87.2\% fewer parameters than SegFormer, and 97.7\% fewer parameters than CPF-Net. This also demonstrates the potential of our proposed approach in optimizing model integration and deployment.

\end{sloppypar}

\printcredits

\section*{Declaration of competing interest}
The authors declare that they have no known competing financial interests or personal relationships that could have appeared to influence the work reported in this paper.

\section*{Declaration of Generative AI and AI-assisted technologies in the writing process}
During the preparation of this work, the author(s) did not use any generative artificial intelligence (AI) or AI-assisted technologies for the writing process. The content was produced solely by the human author(s) without the aid of such tools, and the author(s) take(s) full responsibility for the content of the publication.

\section*{Acknowledgements}
This research is partially supported by the National Key Research and Development Program of China with Grant ID 2018AAA0103203.

\bibliographystyle{elsarticle-num-names}

\bibliography{cas-refs}

\clearpage
\begin{appendices}
\counterwithin{table}{section} 

\section{Evaluation of the different PMFSNet settings}
Conventional models based on the UNet architecture use a decoder to recover the image resolution, especially to minimize the loss of semantic information during up-sampling which is combined with a convolutional operation to extract important features. In our PMFSNet, the traditional 4-stage encoder is reduced to a 3-stage encoder, so the up-sampling step is reduced accordingly. Furthermore, considering that PMFSNet is designed to capture long-term dependencies and is tailored for lightweight segmentation tasks, we have also developed a decoder specifically designed to directly fuse feature maps from multiple scales through a convolution block (see Figure \ref{fig: PMFSNet}). The direct fusion in our design differs from the classic decoder used in the UNet, where up-sampling is progressively achieved through deconvolution. We conduct experiments on various architectures and datasets to examine the impact of the classic decoder in UNet and the direct fusion in our design on performance.

\begin{table}[H]
\centering
\caption{Comparison results of the classic decoder and the direct fusion in the PMFSNet. The best results are in bold. $\uparrow$ means higher values are better.}
\label{tab: decoder ablation}
\setlength{\tabcolsep}{2.8mm}{
    \begin{tabular}{ccc}
    \toprule
    Dataset & Up-sampling & IoU(\%)$\uparrow$ \\
    \midrule
    \multirow{2}{*}{3D CBCT tooth} & direct fusion & \textbf{84.68} \\
       & progressive up-sampling & 84.41 \\
    \midrule
    \multirow{2}{*}{MMOTU} & direct fusion & 80.03 \\
       & progressive up-sampling & \textbf{82.02} \\
    \midrule
    \multirow{2}{*}{ISIC 2018} & direct fusion & 78.19 \\
       & progressive up-sampling & \textbf{78.82} \\
    \bottomrule
    \end{tabular}}
\end{table}

As shown in Table \ref{tab: decoder ablation}, we evaluate the networks with the classic progressive decoder or the direct fusion on the three datasets. We can observe that the PMFSNet3D architecture achieves better segmentation performance (with IoU of 84.68\%) when using direct fusion up-sampling, while the PMFSNet2D architecture achieves better results (with IoU of 82.02\% and 79.82\% on MMOTU and ISIC 2018 datasets, respectively) on both of the other datasets when using the classic progressive decoder.

For medical image segmentation tasks of various scales and dimensions, we design multiple scaling versions of the PMFSNet architecture, including three scales: BASIC, SMALL, and TINY. BASIC is the basic scaling version, which is different from the other two scaling versions in that it comes with a decoder, more channels, and more dense-feature-stacking units at each stage. Specifically, BASIC configures the internal channels of the PMFS block of 64, the base channels of [24, 48, 64] for each stage, the channels of skip connections of [24, 48, 64] for each stage, and the number of dense-feature-stacking units of [5, 10, 10] for each stage. For the SMALL scaling version, the number of channels in its PMFS block is reduced from 64 to 48, with fewer base channels of [24, 24, 24] for each stage, and the channels of skip connections of [12, 24, 24] for each stage. For the smallest scaling version, TINY, the number of dense-feature-stacking units per stage is further reduced to [3, 5, 5]. We quantitatively evaluate their most suitable scaling version on three datasets.

\begin{table}[H]
\centering
\caption{Comparison results of different scaling versions of the PMFSNet. The best results are in bold. $\uparrow$ means higher values are better, $\downarrow$ means lower values are better.}
\label{tab: scaling comparison}
\setlength{\tabcolsep}{0.8mm}{
    \begin{tabular}{ccccc}
    \toprule
    Dataset & Scale & Params(M)$\downarrow$ & HD(mm)$\downarrow$ & IoU(\%)$\uparrow$ \\
    \midrule
    \multirow{3}{*}{3D CBCT tooth} & BASIC & 2.27 & 8.09 & \textbf{86.56} \\
       & SMALL & 1.21 & 12.90 & 85.15 \\
       & TINY & \textbf{0.63} & \textbf{5.57} & 84.68 \\
    \midrule
    \multirow{3}{*}{MMOTU} & BASIC & 0.99 & - & \textbf{82.02} \\
       & SMALL & 0.54 & - & 79.37 \\
       & TINY & \textbf{0.33} & - & 77.72 \\
    \midrule
    \multirow{3}{*}{ISIC 2018} & BASIC & 0.99 & - & \textbf{78.82} \\
       & SMALL & 0.54 & - & 77.82 \\
       & TINY & \textbf{0.33} & - & 77.25 \\
    \bottomrule
    \end{tabular}}
\end{table}

As shown in Table \ref{tab: scaling comparison}. The model baseline we used on the 3D CBCT tooth dataset is PMFSNet3D, whose Parameters (Params) for the scaling versions of BASIC, SMALL, and TINY are 2.27 M, 1.21 M, and 0.63 M, respectively. BASIC is 3.6 times as large as TINY, and SMALL is twice as large as TINY. Regarding segmentation performance, their HDs are 8.09 mm, 12.9 mm, and 5.57 mm, respectively, which suggests that the more lightweight TINY is more suitable for segmentation of surface boundary for our task. BASIC has better segmentation accuracy with the IoU of 86.56\%, outperforming SMALL and TINY by 1.41\% and 1.88\%, respectively. Combined with the observation from Table \ref{tab: 3D CBCT Tooth Comparison}, we believe that the TINY scaling version already achieves optimal performance among competitors, and to balance performance and efficiency, we default to the PMFSNet3D with the TINY architecture. On the MMOTU and ISIC 2018 datasets, we use the PMFSNet2D benchmark architecture. The Parameters (Params) of the three scaling versions, BASIC, SMALL, and TINY, are 0.99 M, 0.54 M, and 0.33 M, respectively, which are all $\leq$ 1 million. The comparative analysis shows that BASIC's IoU on the MMOTU dataset exceeds SMALL and TINY by 2.65\% and 4.3\%, respectively. BASIC's IoU on the ISIC 2018 dataset exceeds SMALL and TINY by 1\% and 1.57\%, respectively. It can be seen that BASIC has a large improvement over the other two scaling versions, and we preferred BASIC for our experiments when the parameters of all three scaling versions are not very large.

\section{Evaluation of the PMFS block channels}
Our proposed PMFS block is a plug-and-play multi-scale feature enhancement module that can be easily extended to the network's bottleneck in any UNet-based architecture. Because of this, we can arbitrarily set the number of channels inside the PMFS block according to the complexity of the networks. To exploit the performance of the PMFS block as much as possible in our experiments, we evaluate the performance of the PMFS block with internal channels of 32, 48, and 64, respectively on multiple scaling versions of the PMFSNet.

\begin{table}[H]
\centering
\caption{Comparison results of different numbers of channels in the PMFS block. The best results are in bold. $\uparrow$ means higher values are better, $\downarrow$ means lower values are better.}
\label{tab: channel comparison}
\setlength{\tabcolsep}{2.1mm}{
    \begin{tabular}{ccccc}
    \toprule
    Dataset\&Scale & Channel & Params(M)$\downarrow$ & IoU(\%)$\uparrow$ \\
    \midrule
    \multirow{3}{*}{Tooth\&3D-TINY} & 32 & \textbf{0.54} & 84.32 \\
       & 48 & 0.63 & \textbf{84.68} \\
       & 64 & 0.76 & 84.41 \\
    \midrule
    \multirow{3}{*}{MMOTU\&2D-BASIC} & 32 & \textbf{0.76} & 80.66 \\
       & 48 & 0.85 & 80.58 \\
       & 64 & 0.99 & \textbf{82.02} \\
    \midrule
    \multirow{3}{*}{ISIC 2018\&2D-BASIC} & 32 & \textbf{0.76} & 78.44 \\
       & 48 & 0.85 & 78.37 \\
       & 64 & 0.99 & \textbf{78.82} \\
    \bottomrule
    \end{tabular}}
\end{table}

The evaluation results are shown in Table \ref{tab: channel comparison}, which shows that the Parameters (Params) of the networks do not cause large fluctuation when the internal channels of the PMFS block are set to 32, 48, and 64, respectively. The PMFSNet3D-TINY network achieves the best performance on the 3D CBCT tooth dataset when the PMFS block internal channel is set to 48, with the IoU of 84.68\%, so for the TINY scaling version of the PMFSNet we set the internal channel of the PMFS block to 48. The PMFSNet2D-BASIC network on the MMOTU dataset achieves the best performance when the internal channel of the PMFS block is set to 64, and the IoU is 1.36\% and 1.44\% higher than when the channel setting is 32 and 48, respectively. Similarly, the highest IoU of 78.82\% is achieved on the ISIC 2018 dataset.
\end{appendices}

\end{document}